\documentclass{article}

\usepackage[preprint, nonatbib]{neurips_2026}
\usepackage[numbers, compress]{natbib}

\DeclareUnicodeCharacter{2113}{\ensuremath{\ell}}
\DeclareUnicodeCharacter{03B1}{\ensuremath{\alpha}}

\usepackage[utf8]{inputenc} 
\usepackage[T1]{fontenc}    
\PassOptionsToPackage{hyphens}{url} 
\usepackage{hyperref}       
\usepackage{url}            
\usepackage{booktabs}       
\usepackage{amsfonts}       
\usepackage{nicefrac}       
\usepackage{microtype}
\usepackage{xcolor}
\usepackage{amsmath} 
\usepackage{mathtools}
\usepackage{wrapfig}
\usepackage{float}
\usepackage{subcaption}
\usepackage{array}
\usepackage{moresize}
\usepackage{anyfontsize} 
\usepackage{enumitem}
\usepackage{dsfont}

\title{Resolving superposition in AI for
interpretability and cross-modal alignment in patient-neuronal images}

\author{
  \textbf{Jisung Park}$^1$, 
  \textbf{Seohyeon Kang}$^1$, 
  \textbf{Daeun Yoo}$^1$, 
  \textbf{Eunsu Lee}$^2$, 
  \textbf{Seoin Cho}$^1$, 
  \textbf{Wooyeop Choi}$^1$ \\ [0.1cm]
  \textbf{Ian Choi}$^1$, 
  \textbf{James R. Evan}$^4$,
  \textbf{Daesoo Kim}$^{1,3}$, 
  \textbf{Sonia Gandhi}$^4$, 
  \textbf{Minee L. Choi}$^{1,3}$\thanks{Corresponding author. Email: \texttt{minee.choi@kaist.ac.kr}} \\ [0.25cm]
  $^1$KAIST, \ 
  $^2$Konyang University, \
  $^3$Chang Gung University, \\ [0.05cm]
  $^4$UCL Queen Square Institute of Neurology \& The Francis Crick Institute
}

\newcommand{\citeurl}[1]{} 
\AtBeginDocument{\let\url\citeurl}
\begin{document}

\setlength{\abovedisplayskip}{3pt plus 1pt minus 1pt}

\maketitle

\begin{abstract} 
Artificial intelligence is transforming our capability to solve biological challenges. In dimensionality bottleneck regimes exacerbated by high-dimensional biological data, neural networks force distinct concepts into the lower dimensions known as superposition. Although this superposition is widely known to hinder interpretability, its impact on corrupting the geometry of latent spaces remains critically overlooked. Here, we utilized sparse autoencoders (SAEs) trained on over 100,000 multiplexed images of patient-derived Parkinson's disease and healthy neurons to resolve superposition. This approach bypasses the mathematical non-uniqueness of feature attribution by shifting to interpretable latent representation analysis. We theoretically and empirically demonstrate that superposition contaminates representational metric spaces, and thereby SAEs successfully recover geometric fidelity. By treating these geometrically purified representations as single-cell state vectors, we adapted single-cell RNA sequencing (scRNA-seq) data analysis methodologies directly to the image domain. Finally, we introduce GW-map, utilizing Gromov-Wasserstein optimal transport to align these image representations with authentic scRNA-seq data \emph{de novo}. This coupling reconstructs hierarchical neuronal pathology pathways such as Calcium-AIS scaffold, without reference spatial transcriptomics, establishing a scalable foundation for spatial biology. Code is available at https://github.com/jijihihi/Bio\_superposition

\end{abstract}

\section{Introduction and Related Work}
\subsection{Limitation of Feature Attribution XAI}
Deep learning has become indispensable in biological image analysis, with pathology foundation models routinely extracting high-dimensional representations from visual information. To establish model reliability and generate biological hypotheses, post-hoc feature attribution techniques such as Gradient-weighted Class Activation Mapping (Grad-CAM) \cite{selvarajuGradCAMVisualExplanations2017} and Shapley Additive Explanations (SHAP) \cite{lundbergUnifiedApproachInterpreting2017} are widely employed \cite{aravindkumarExplainableAIHealthcare2026, zhouXAIMeetsBiology2023}. However, these methods confront a fundamental tension rooted in geometry. Under the manifold hypothesis, component-wise credit assignment becomes mathematically non-unique \cite{sundararajanManyShapleyValues2020}. This ambiguity is severely exacerbated in biology due to the dense interdependency of biological variables. Conversely, bypassing this ambiguity by ignoring the data’s intrinsic geometry yields off-manifold explanations \cite{fryeAsymmetricShapleyValues2020, andersFairwashingExplanationsOffmanifold2020}. Given this manifold-attribution paradox, compounded by frequent inconsistencies across explainable AI (XAI) methods \cite{krishnaDisagreementProblemExplainable2024} and other limitations \cite{bilodeauImpossibilityTheoremsFeature2024, huangFailingsShapleyValues2024}, the application of XAI in biology urgently requires an intrinsic alternative.

\subsection{Mechanistic Interpretability and Representation Contamination Exacerbated in Biology}
Mechanistic Interpretability (MI) has recently emerged as a powerful alternative \cite{elhageMathematicalFrameworkTransformer2021}. MI posits that interpretability challenges stem from concept superposition: the forced compression of multiple distinct concepts into fewer dimensions in a neural network \cite{elhageToyModelsSuperposition2022b}. To resolve this, MI utilizes sparse Autoencoders (SAEs) to disentangle superposed concepts into monosemantic units \cite{brickenMonosemanticityDecomposingLanguage, templetonScalingMonosemanticityExtracting}. This conceptual isolation renders the model’s internal representations directly interpretable, obviating manifold-attribution paradox. It achieves human-centric interpretability without sacrificing predictive capacity, altering model architectures \cite{chenThisLooksThat2019a} or predefined concepts \cite{kohConceptBottleneckModels2020, kimInterpretabilityFeatureAttribution2018}.
\enlargethispage{15pt}
While the impact of superposition on interpretability is increasingly recognized \cite{somvanshiBridgingBlackBox02weol42026}, its role in corrupting representation geometry has been critically overlooked \cite{elhageToyModelsSuperposition2022b, brickenMonosemanticityDecomposingLanguage, longonSuperpositionDisentanglementNeural2026}. Because neural networks densely pack concept vectors to optimize loss \cite{elhageToyModelsSuperposition2022b, parkLinearRepresentationHypothesis07weol212024}, this packing contaminates the metric space and destroys the mathematical isometry of the latent space rendering local measurements (e.g., cosine distances) unreliable. 

Crucially, superposition becomes extraordinarily prominent in the biological domain. The internal dimensions of neural networks are vastly narrower than the extreme diversity of biological states, creating severe dimensionality constraints. Furthermore, because interference fundamentally arises from inner products between sparse concept vectors, data sparsity is key factor in superposition. Given that extreme high-dimensionality and zero-inflated sparsity are inherent to biological data \cite{jiangStatisticsBiologyZeroinflation2022}, this geometric corruption is profoundly pronounced in biological AI, fundamentally threatening the reliability of prevalent representation-based downstream analyses \cite{chaiOpportunitiesChallengesDeep2024, caicedoWeaklySupervisedLearning2018a}.

\vspace{-3pt}
\subsection{Our Proposed Framework}
To address this, we employed SAEs to disentangle the superposition. We validated this framework using patient induced pluripotent stem cell (iPSC)-derived cortical neurons in a high-content assay. Through theoretical and empirical analysis, we demonstrate that superposition contaminates representation and thereby disentangling superposition and lifting the dimensionality via SAEs restores the corrupted metric space, empowering the representations to faithfully reflect intrinsic data semantics. Leveraging these geometrically purified representations and structural similarity between SAE representation and single-cell RNA sequencing (scRNA-seq) data (see section 5), we successfully deployed graph-based scRNA-seq analytical methods tailored to the intrinsic noise and continuous biological properties of the data directly to the image domain. 

Furthermore, we developed the GW-map framework utilizing Gromov-Wasserstein (GW) optimal transport \cite{peyreComputationalOptimalTransport02weol122019, memoliGromovWassersteinDistances2011, demetciSCOTSingleCellMultiOmics2022} to \emph{de novo} align image and single-cell transcriptomic data without reference spatial transcriptomics (ST). By autonomously reconstructing the hierarchical pathway architecture of neuronal pathology, this approach overcomes the fundamental limitations of ST, including restricted cellular resolution, high costs, limited gene coverage, and chronic cross-modal alignment challenges \cite{guoIntegratingSpatiallyResolvedTranscriptomics2025, khanComprehensiveReviewSpatial2025}.

The main contributions of this paper are summarized as follows:
\begin{itemize}[noitemsep, topsep=1pt, parsep=0pt, partopsep=0pt]
    \item We theoretically demonstrate that superposition mathematically contaminates the representation metric space.
    \item We empirically demonstrate this geometric contamination in biological imaging models and show that SAE-based superposition disentanglement faithfully recovers the intrinsic geometric fidelity of the data
    \item Leveraging geometrically purified SAE representations, we successfully adapted advanced scRNA-seq analysis algorithms to the image domain, establishing a powerful framework for evaluating representations and uncovering biological hypotheses.
    \item We introduce GW-map, which autonomously couples purified SAE representations with authentic scRNA-seq data via Gromov-Wasserstein optimal transport, suggesting a highly scalable new approach to spatial biology
\end{itemize}

\section{Method}
In this section, we present the computational framework that transforms high-content imaging data into interpretable, geometrically purified representations, adapts of scRNA-seq methodologies to the image domain, and aligns with authentic scRNA-seq data via GW coupling. All experimental results are reported as the average over eight independent random seeds. Detiled methods in Appendix E.

\begin{figure}[H]
    \vspace{-20pt}
    \centering
    \includegraphics[width=0.95\linewidth]{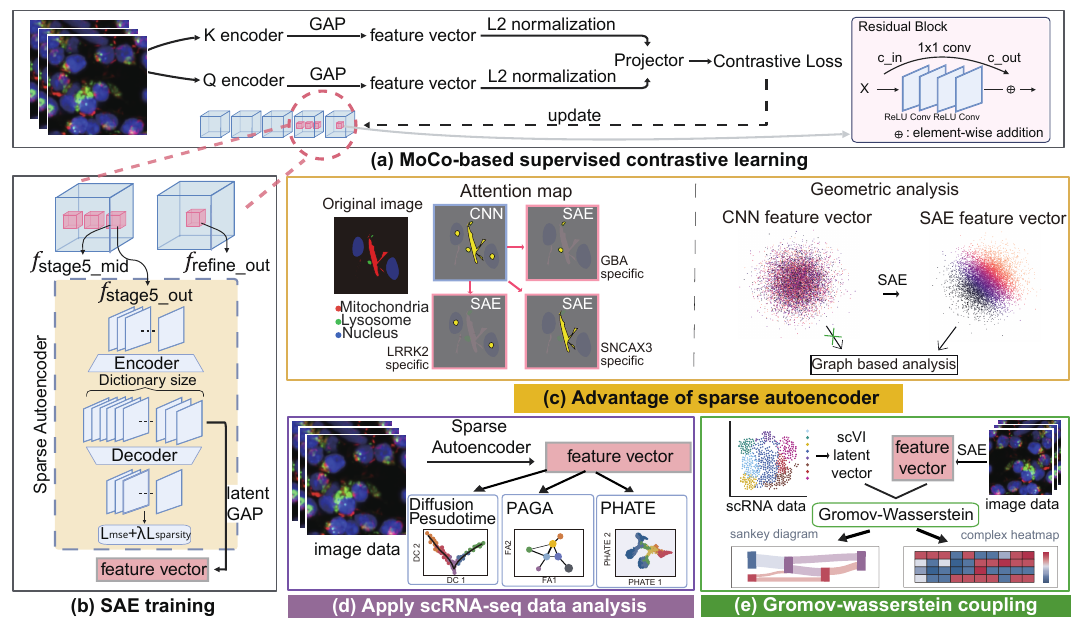}
    \caption{\textbf{(a)} MoCo-based contrastive representation
learning framework. \textbf{(b)} Integration of the SAE to CNN. SAE GAP latent vectors were used for pseudo-transcriptomic analysis and multimodal integration. \textbf{(c)} Left: Comparison between entangled CNN attention and disentangled attention of the SAE. Right: SAE more faithfully reflects the data topological relationships. \textbf{(d)} Single-cell methods were applied to SAE representations. \textbf{(e)} Unsupervised cross-modal alignment of SAE representations with authentic scRNA-seq data. Downstream analyses validated the biological plausiblity of this cross-modal alignment. Panels (a-e) were created using BioRender; Choi, ML. (2026) https://BioRender.com/tk1hkpy}
    \vspace{-8pt}
    \label{FIG1}
\end{figure}

\subsection{Contrastive Learning of CNN on Cortical Neuronal Images}
Cortical neurons were differentiated from hiPSCs of healthy controls (3 lines) and familial PD patients \emph{(SNCA $\times 3$, LRRK2, GBA)}. High-content live-cell imaging was performed using four fluorescent markers: Hoechst 33342 (nuclei), TMRM (mitochondria), LysoTracker (lysosomes), and Sytox Green (dead cells for apoptosis quantification) \cite{dsaPredictionMechanisticSubtypes2023}. A custom CNN architecture was chosen over a Vision Transformer (ViT) due to its spatial locality \cite{lecunGradientbasedLearningApplied1998, olahZoomIntroductionCircuits2020, burgertImageNettrainedCNNsAre2025a, brendelApproximatingCNNsBagoflocalFeatures2019a, dosovitskiyImageWorth16x162021, raghuVisionTransformersSee2021}, aiding mechanistic interpretation. The network was trained on over 100,000 images using a three-channel input (excluding the dead-cell marker). Figure 1 (a) shows the supervised contrastive learning framework \cite{khoslaSupervisedContrastiveLearning2020} based on MoCo \cite{heMomentumContrastUnsupervised2020a} for four-class classification (control and three PD mutations). To minimize technical artifacts and variations in cell density, instance normalization (IN) was applied to all input images.

\subsection{Sparse Autoencoder Integration and Representation Evaluation}
To identify the optimal latent space, we evaluated feature vectors extracted via Global Average Pooling (GAP) from the final three CNN layers preceding the projection head. Gated sparse autoencoders (SAE) \cite{brickenMonosemanticityDecomposingLanguage, templetonScalingMonosemanticityExtracting, rajamanoharanImprovingSparseDecomposition2024} were trained on point-wise vectors ($v_{ij} \in\ R^C$)  from the spatial locations of the optimal CNN feature map ($f_{stage5} \in \mathbb{R}^{H\times W\times C}$) (Fig. 1(b)). Following standard practice, each point-wise vector was batch-centered $L_2$-normalized prior to SAE training. We evaluated superposition disentanglement using inner product distributions, spatial attention monosemanticity and effective dimensionality (eRank) across various expansion factors and sparsity penalties. The SAE successfully decomposed complex spatial attention into human-interpretable feature maps (Fig. 1(c), left) and preserved biological semantics with significantly higher topological fidelity than the baseline CNN representations. (Fig. 1(c), right). A sparsity threshold of $1\times{10}^{-5}$ was applied to the SAE activations for all subsequent analyses unless otherwise stated.

\subsection{Manifold Analysis via Adaptation of scRNA-seq Methodologies to Image Domain}
Leveraging the and semantically resolved features from the SAE, we treated these representations as transcriptomic-like single-cell profiles. This conceptual bridge enabled the direct adaptation of scRNA-seq analytical frameworks, which offer the tools for non-linear manifold analysis. Specifically, we utilized Diffusion Pseudotime (DPT) \cite{haghverdiDiffusionPseudotimeRobustly2016}, which approximates geodesic distances across complex, non-Euclidean data manifolds by modeling stochastic random walks and heat diffusion kernels. We also applied the Potential of Heat-diffusion for Affinity-based Transition Embedding (PHATE) \cite{moonVisualizingStructureTransitions2019a} to visualize information-theoretic distances, and Partition-based Graph Abstraction (PAGA) \cite{wolfPAGAGraphAbstraction2019} to rigorously quantify topological connectivity.


\subsection{Cross-Modal Coupling via GW-map}
We coupled image and authentic scRNA-seq via GW-map. Specifically, we batch-corrected publicly available \emph{SNCA $\times 3$} 3D organoid scRNA-seq data \cite{jinModelingLewyBody2024} using single-cell variational inference (scVI) \cite{lopezDeepGenerativeModeling2018} and computationally filtered them to retain 2D-representative populations. The PCA-reduced SAE feature vectors were then structurally aligned with the scVI latent vectors via GW coupling. To robustly quantify gene-specific predictability without variance-induced inflation, we calculated $Z$-scores by standardizing the predictive $R^2$ values against a null distribution generated via rigorous permutation testing. Downstream analyses validated the biological plausibility of this cross-modal alignment.

\vspace{-5pt}
\section{Biological Information Encoded in the CNN Representation}
\vspace{-5pt}

Our custom CNN achieved a linear probe accuracy of 90.31 ± 0.38\% (mean ± SD; Fig. 2(a), (b)), and Centered Kernel Alignment (CKA) \cite{kornblithSimilarityNeuralNetwork2019a} confirmed robust representational similarity across independently trained models (512 dimensions, 2,500 samples/class \cite{nguyenWideDeepNetworks2021a, murphyCorrectingBiasedCentered2024}; Fig. 2(c)). To evaluate whether the network captures biological information beyond classification, we predicted cell death rates using representations from the final three CNN layers ($f_{stage5_mid}$, $f_{stage5_out}$, and $f_{refine_out}$) via ridge regression \cite{chenXGBoostScalableTree2016}. Although the model was trained exclusively on organelle markers without cell death labels, all three layers significantly predicted cell death rates across mutation lines (Fig. 2 (d)).

Notably, $f_{stage5_out}$ yielded the highest predictive performance (cross-validation $R^2$ up to 0.52 for \emph{SNCA $\times 3$}, permutation $p < 0.01$), making it the optimal input layer for the subsequent SAE training. In contrast, the final $f_{refine_out}$ layer exhibited dimensional collapse \cite{papyanPrevalenceNeuralCollapse2020}, marked by sharp declines in both predictive $R^2$ and eRank \cite{royEffectiveRankMeasure2007} (Fig. 2(d), (e)). UMAP projections of $f_{stage5_out}$ revealed distinct class-specific topologies, where the implicitly encoded biological information manifested as a continuous gradient of cell death (Fig. 2(f), (h). This robust prediction of \emph{SNCA $\times 3$}-associated cell death aligns with established $\alpha$-synuclein pathology in Parkinson's disease\cite{singletonAlphaSynucleinLocusTriplication2003, chartier-harlinAlphasynucleinLocusDuplication2004, ludtmannAsynucleinOligomersInteract2018} Finally, because CNN feature magnitudes are susceptible to spatial artifacts such as cell density variations, we applied $L_2$-normalization to all feature vectors prior to subsequent analyses (Justifications in Appendix A).

\enlargethispage{1\baselineskip}

\begin{figure}[H]
    \centering
    \includegraphics[width=0.9\linewidth]{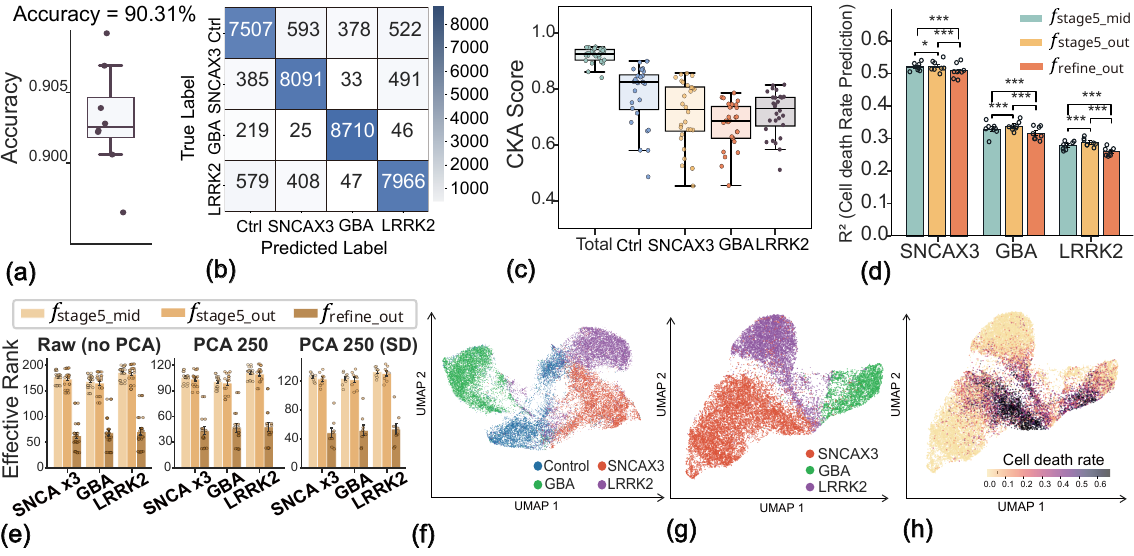}
    \vspace{-4pt}
    \captionof{figure}{\textbf{(a, b)} Linear classification performance and confusion matrix of CNN. \textbf{(c)} CKA representational similarity between independent models showing robust convergence. \textbf{(d--e)} Ridge regression prediction of cell death rates and effective rank analysis using the final three CNN layer representations. \textbf{(f--h)} UMAP projection of CNN $f_{\text{stage5\_out}}$. Panels \textbf{(f--g)} are color-coded by class, showing mutation classes in \textbf{(g--h)}, while \textbf{(h)} is colored by cell death rate. Statistical significance in (d) was assessed using a two-sided Wilcoxon signed-rank test, with each cross-validation fold treated as an independent observation. (*p < 0.05, **p < 0.01, ***p < 0.001)}
    \vspace{-15pt}

\end{figure}
\clearpage

\section{Disentanglement of Superposition: Interpretability and Geometric Fidelity}
\subsection{SAE Superposition Disentanglement}

To investigate the resolution of concept superposition, we swept Gated SAE dictionary sizes ($d$) and sparsity penalties ($\lambda$) (Table 1). We observed a clear Pareto trade-off between reconstruction fidelity (FVU), linear probe accuracy, and $L_0$ sparsity. As a baseline proxy for the CNN, low-dimensional, low-sparsity SAE was trained. Scaling $d$ and $\lambda$ generally decreased pairwise cosine similarities (inner products) between concept vectors while increasing the number of active feature maps, confirming the unraveling of dense CNN entanglement.

\begin{table}[h]
\vspace{-3pt}

\renewcommand{\arraystretch}{0.85}
\centering

\caption{Evaluation of Gated SAE scaling behavior across dictionary sizes ($d$) and sparsity penalties ($\lambda$). Alive Feat. denotes the number of active feature maps.}
\label{tab:sae_evaluation}

\fontsize{9pt}{11pt}\selectfont 
\setlength{\tabcolsep}{4.5pt}

\begin{tabular}{
  >{\centering\arraybackslash}p{\dimexpr 0.135\textwidth - 2\tabcolsep} 
  >{\centering\arraybackslash}p{\dimexpr 0.120\textwidth - 2\tabcolsep} 
  >{\centering\arraybackslash}p{\dimexpr 0.060\textwidth - 2\tabcolsep} 
  >{\centering\arraybackslash}p{\dimexpr 0.075\textwidth - 2\tabcolsep} 
  >{\centering\arraybackslash}p{\dimexpr 0.185\textwidth - 2\tabcolsep} 
  >{\centering\arraybackslash}p{\dimexpr 0.080\textwidth - 2\tabcolsep} 
  >{\centering\arraybackslash}p{\dimexpr 0.158\textwidth - 2\tabcolsep} 
  >{\centering\arraybackslash}p{\dimexpr 0.158\textwidth - 2\tabcolsep} 
}
\toprule
\textbf{Model} & \textbf{Dimension} & $\lambda$ & \textbf{\boldmath{$L_0$}} & \textbf{Alive Feat.} & \textbf{FVU} & \textbf{Accuracy} & \textbf{Inner Product} \\
\midrule
CNN            & 512  & N/A & N/A & 512  & N/A  & 90.31 & N/A \\
CNN Proxy      & 600  & 50  & 4.9 & 587  & 0.34 & 89.24 & 0.160 \\

\midrule 

Gated SAE      & 1024 & 800 & 2.7 & 1021 & 0.73 & 88.48 & 0.104 \\
Gated SAE      & 2048 & 800 & 4.2 & 2035 & 0.53 & 89.63 & 0.089 \\
Gated SAE      & 4096 & 800 & 7.6 & 3970 & 0.40 & 89.52 & 0.082 \\
Gated SAE      & 8192 & 800 & 9.8 & 3733 & 0.30 & 89.91 & 0.104 \\

\midrule 

Gated SAE      & 4096 & 200  & 13.2 & 3850 & 0.24 & 89.52 & 0.099 \\
Gated SAE      & 4096 & 800  & 7.6  & 3970 & 0.40 & 89.88 & 0.082 \\
Gated SAE      & 4096 & 3200 & 2.9  & 3866 & 0.72 & 88.23 & 0.052 \\
\bottomrule
\end{tabular}
\vspace{-10pt}
\end{table}

\begin{figure}[h]
  \centering
  \nopagebreak
  \includegraphics[width=1.0\linewidth]{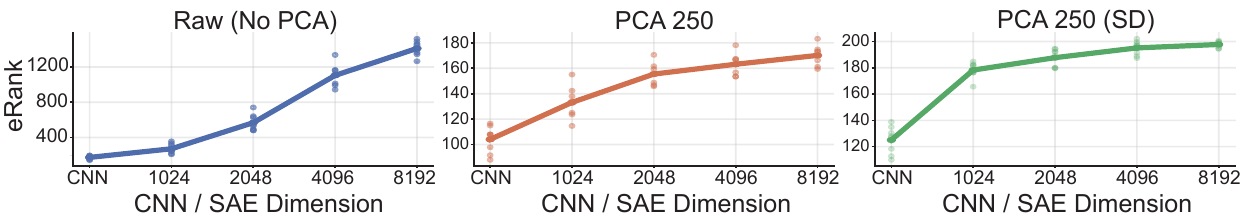}
  \captionof{figure}{The eRank consistently increases from the baseline CNN ($f_{stage5_out}$) as the SAE dictionary size scales under a fixed $\lambda$ regime. This trend is robustly observed across raw latent vectors, PCA projections, and PCA on standardized features (left to right). SD denotes standard deviation.
  }
  \vspace{-2pt}
  \label{Figure 3}
\end{figure}

Furthermore, increasing $d$ under a fixed $\lambda$ consistently expanded the effective rank (eRank) relative to the CNN baseline across raw and PCA-reduced latent spaces (Fig. 3). Although an increase in inner product magnitude was observed at $d=8192$, increasing $d$ under a fixed $\lambda$ consistently expanded the eRank relative to the CNN baseline. This eRank expansion quantitatively demonstrates that polysemantic concepts, previously packed into overlapping dimensions, are successfully disentangled into independent directions.

\subsection{Intrinsic Interpretability Bypassing the Manifold-Attribution Paradox}

We used an SAE with $\lambda = 3200$ for interpretability analyses. While conventional CNN feature maps polysemantically co-activate across distinct concepts (e.g., simultaneously firing for mitochondria and lysosomes), the SAE effectively isolates these into monosemantic activations representing specific, human-interpretable biological structures (e.g., nuclear envelopes), confirming successful superposition disentanglement (Fig. 4(a)). This result indicates that superposition fundamentally compromises the reliability of spatial aggregation metrics such as GAP. Because polysemantic CNN feature maps collapse distinct biological structures into a single activation vector, high GAP scores cannot be reliably attributed to specific causal features, thereby undermining post-hoc attribution methods like Grad-CAM. Conversely, the monosemanticity achieved by the SAE transforms GAP into a trustworthy quantifier of isolated concepts. Consequently, the SAE yields significantly more class-specific feature maps than the CNN baseline (Fig. 4(b)), establishing a rigorous link between classes and specific visual characteristics. This intrinsic interpretability obviates the need for post-hoc feature attribution, and effectively bypasses the manifold-attribution paradox.

Theoretically, in the dimensionality bottleneck regime, networks pack disparate concepts into the same direction, which severely inflates the local Lipschitz constant. Because many post-hoc XAI methods assume local gradient stability \cite{lundbergUnifiedApproachInterpreting2017, anconaBetterUnderstandingGradientbased2018} (requiring a small Lipschitz constant), heavily superposed networks inherently violate these prerequisites across class boundaries. By resolving feature entanglement, the SAE restores local perturbation additivity, justifying the approximation of neural network units as local additive linear models. Thus, SAE disentanglement not only provides intrinsic interpretability but also satisfies the fundamental structural prerequisites required for traditional XAI methods to function reliably. Consequently, integrating existing XAI methods to this disentangled SAE latent space offers a framework to further improve interpretability.

The biological fidelity of these monosemantic concepts was further validated via Activation Maximization (AM), which consistently aligned with human-interpretable structures such as the perinuclear clustering of mitochondria and lysosomes (Fig. 4(c); additional images in Appendix F.3).

\begin{figure}[t]
  \centering
  \nopagebreak
  \includegraphics[width=1.0\linewidth]{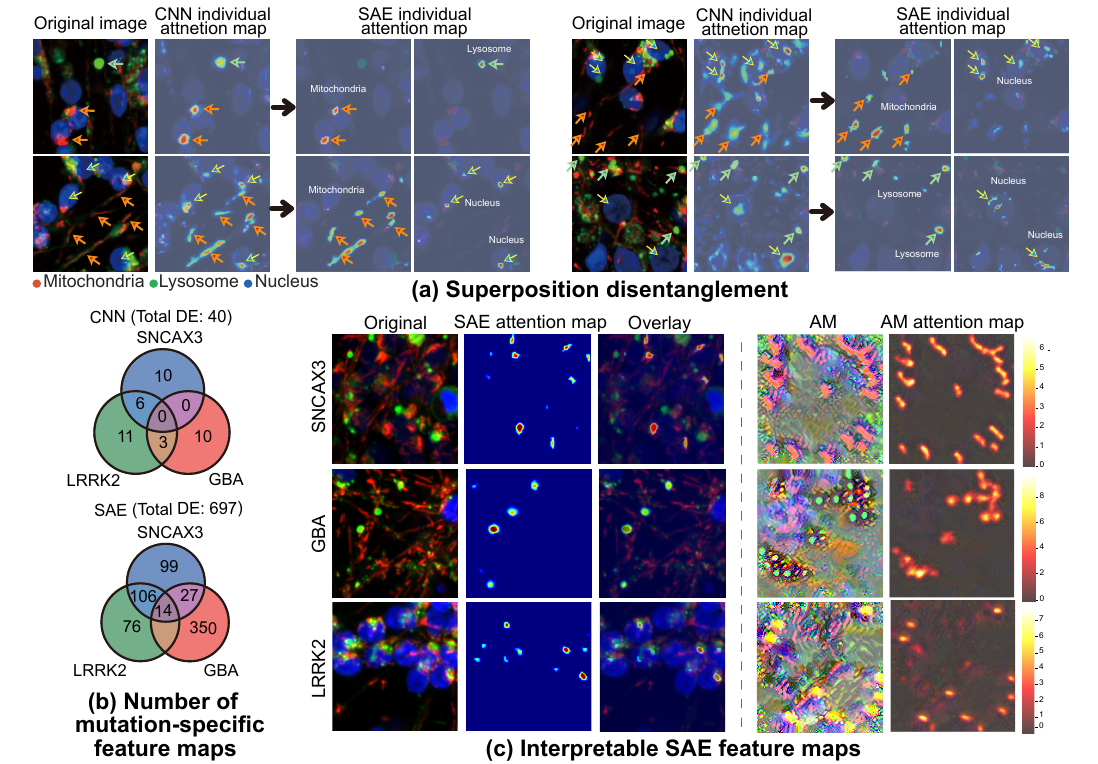}
  \captionof{figure}{\textbf{(a)} Visual disentanglement of superposed concepts in SAE. Representative original images and bilinear-interpolated attention maps. The SAE reconstruction loss was spatially weighted by the $L_2$ norm of point-wise feature vectors $v_{ij} \in\ R^C$ to preserve the original CNN spatial activation magnitudes indicative of local token importance.\textbf{(b)}. Quantification of mutation-specific feature maps across three pairs of healthy to three PD mutation groups via differential activation ($|{log}_2FC| > 0.58$ with sparsity threshold $5\times10^{-5}$.) \textbf{(c)}. Visual validation of mutation-specific SAE concepts, displaying original images, spatial attention maps, and overlays (left), alongside AM images and corresponding attention maps (right)}
  \vspace{-10pt}
  \label{figure_interpretability}
\end{figure}

\subsection{Superposition-Induced Geometric Contamination and Recovery of Semantic Fidelity}

\paragraph{Theoretical Analysis: Geometric Contamination due to Superposition}
In this section, we analyze the representational impact caused by superposition (Details in Appendix B.). According to the Linear Representation Hypothesis \cite{elhageToyModelsSuperposition2022b, parkLinearRepresentationHypothesis07weol212024}, a latent representation $h_i \in \mathbb{R}^d$ is formed through a linear combination of learned concept vectors ${w_1, \dots, w_m} \subset \mathbb{R}^d$, where $m$ is the number of encoded concepts ($m > d$). That is, $h_i = \sum_{k=1}^m v_{i,k} w_k$, where $v_{i,k}$ denotes the semantic activation of concept $k$. This dimensional bottleneck forces neural networks to strategically tolerate superposition, compressing distributed features into overlapping directions to optimize the loss \cite{troppGreedGoodAlgorithmic2004, livezeyLearningOvercompleteLow2019}. This systemic geometric packing stress remains a critical bottleneck.
\enlargethispage{1\baselineskip}
In real-world data, intrinsic concepts are fundamentally non-independent, exhibiting dense semantic correlations. To capture this context-dependent geometry, we model the representations as residing on a non-linear Riemannian manifold $\mathcal{M}$. For any given semantic context $p \in \mathcal{M}$, the unconstrained semantic proximity of concepts is governed by the local target Gram matrix $\mathbf{G}_p \in \mathbb{R}^{m \times m}$ induced by the Riemannian metric tensor $g_p$. Let $\mathbf{W} = [w_1, \dots, w_m] \in \mathbb{R}^{d \times m}$ be the dictionary matrix corresponding to the network weights (or SAE decoder). Under this localized framework, embedding these correlated concepts into a strictly lower $d$-dimensional space is limited by the generalized Welch bound \cite{welchLowerBoundsMaximum1974, waldronGeneralizedWelchBound2003}, which imposes a strict algebraic lower bound on the overall metric distortion error tensor $\mathbf{E}_p(d) = \mathbf{W}^T\mathbf{W} - \mathbf{G}_p$:

\begin{equation} \|\mathbf{E}_p(d)\|_F^2 = \|\mathbf{W}^T\mathbf{W} - \mathbf{G}_p\|_F^2 \geq \frac{(\text{Tr}(\mathbf{G}_p) - d)^2}{m - d} \end{equation}

This strictly positive lower bound mathematically guarantees that the empirical dictionary geometry $\mathbf{W}^T\mathbf{W}$ cannot be free from packing artifacts \cite{troppGreedGoodAlgorithmic2004}. In our generalized non-independent regime, this constraint forces $\mathbf{W}$ towards a deformed, generalized ETF-like configuration \cite{papyanPrevalenceNeuralCollapse2020, jiangGeneralizedNeuralCollapse07weol212024}. Consequently, the empirical inner product inherits this geometric deformation: $w_k^T w_l = (\mathbf{G}p){k,l} + (\mathbf{E}p){k,l}(d)$.

This directly pollutes downstream representations. The empirical inner product between any two latent vectors $h_i$ and $h_j$ within the local tangent space $T_p\mathcal{M}$ expands into:
\begin{equation}
    h_i^T h_j = \underbrace{\sum_{k=1}^m v_{i,k} v_{j,k}(\mathbf{G}_p)_{k,k} + \sum_{k \neq l} v_{i,k} v_{j,l} (\mathbf{G}_p)_{k,l}}_{\text{Pure Local Semantic Component ($S_{ij}$)}} + \underbrace{\sum_{k \neq l} v_{i,k} v_{j,l} (\mathbf{E}_p)_{k,l}(d)}_{\text{Coupled Geometric Contamination ($G_{ij}$)}}
\end{equation}

Since $\|\mathbf{E}_p(d)\|F^2 > 0$ strictly holds by Eq. (1), the geometric contamination $G_{ij}$ cannot be nullified in a bottlenecked model. Because the $L_2$ metric is derived from the inner product, this directly corrupts the Euclidean distances. Within $T_p\mathcal{M}$, this $L_2$ packing distortion propagates across all local $L_p$ spaces ($p \geq 1$) via the equivalence of norms ($C_1 |h|_2 \leq |h|_p \leq C_2 |h|_2$). Consequently, latent Euclidean distances fail to purely reflect true semantic distances, fundamentally corrupted by the geometric packing stress of the overcomplete mapping. Furthermore, assuming semantic data resides on a non-linear manifold, dimensional bottlenecks structurally preclude exact isometric embeddings, extending this geometric contamination to arbitrary metric spaces.

SAEs circumvent this structural limitation by projecting into a higher-dimensional space to unpack superposed concepts, thereby resolving the rank deficiency and recovering a metric space that faithfully reflects the underlying semantic similarities. This dimensional expansion, coupled with the extreme activation sparsity of SAEs, restores orthogonality by providing sufficient white space to mathematically decouple the superposed concepts (driving $\|\mathbf{E}_p(d)\|_F^2 \rightarrow 0$). Consequently, from a theoretical standpoint, the disentangled metric space of SAEs is semantically far more reliable than the geometrically contaminated latent spaces of dimensionally bottlenecked models.

\paragraph{Empirical Validation of Geometric Contamination in Bottlenecked Representations}

\begin{figure}[t]
  \centering
  \nopagebreak
  \includegraphics[width=1.0\linewidth]{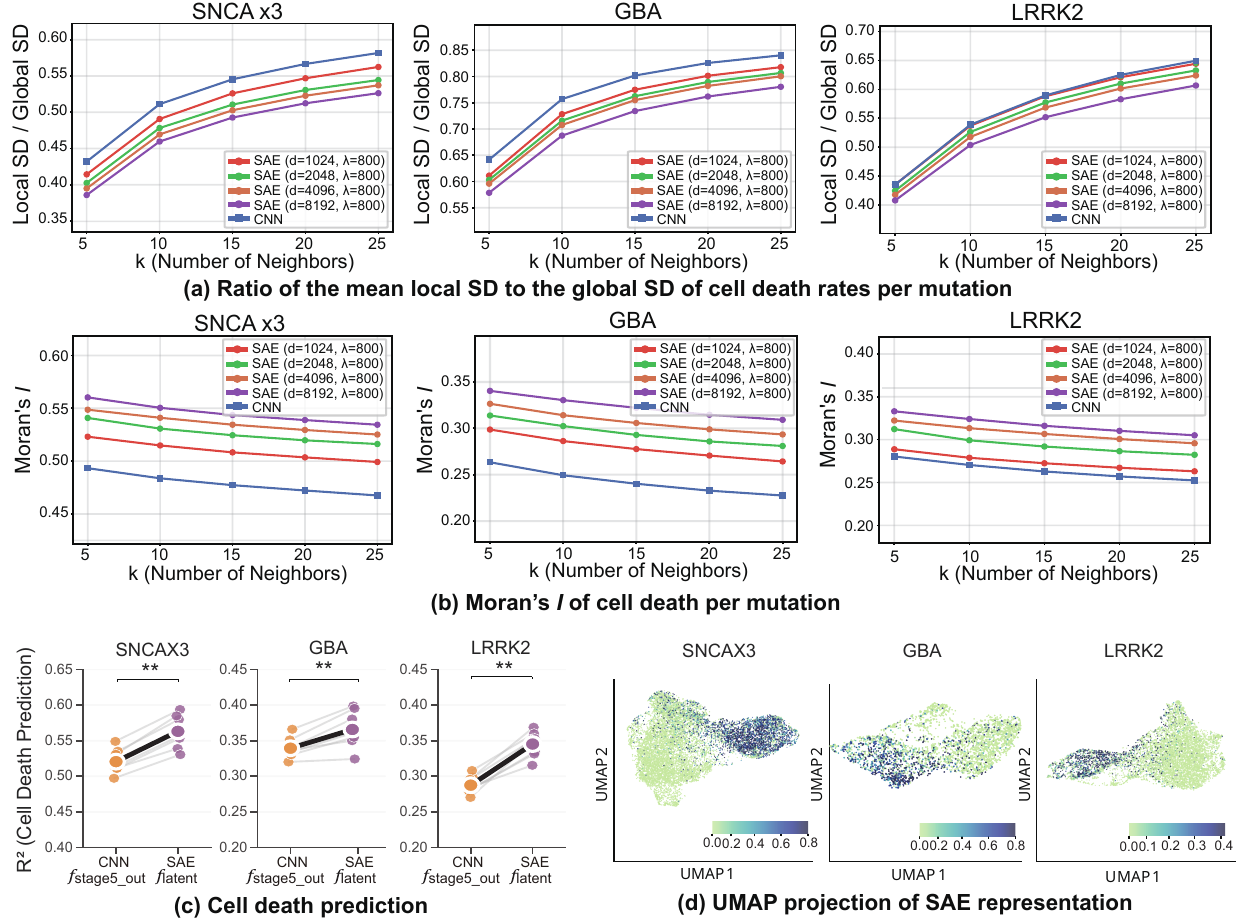}
  \captionof{figure}{Empirical validation of geometric contamination and its recovery. \textbf{(a)} Quantification of the local semantic dispersion metric across individual mutation lines to account for confounding effects. \textbf{(b)} Moran's $I$ are evaluated for each mutation group, exposing suppressed spatial consistency in the CNN baseline. \textbf{(c)} Predictive performance (cross-validation $R^2$) of cell death rates via ridge regression \textbf{(d)} UMAP projection of the SAE latent space, colored by the continuous cell death rate.}

  \vspace{-10pt}
  \label{knn_std_cellpredict_combination}
\end{figure}

To empirically demonstrate that bottlenecked representations suffer from geometric contamination , we leveraged continuous, image-specific cell death rates, an entirely unseen biological state. If a latent space is geometrically uncorrupted, semantic neighbors should naturally exhibit highly similar functional states. By employing the SAE to systematically relieve dimensional packing stress, we reveal this underlying distortion.

We quantified this geometric distortion using a local semantic dispersion metric: the ratio of the k-nearest neighbors (kNN) cell death SD to each mutation-specific SD Across all $k$ values, the baseline CNN $f_{stage5_out}$ exhibited significant semantic dispersion. Conversely, relieving the dimensional bottleneck via SAE significantly reduced this ratio across all mutation lines (Fig. 5(a)). Scaling the SAE dictionary size systematically mitigated this metric, confirming that the initial semantic distortion in the CNN was indeed an artifact of geometric packing. Spatial autocorrelation analysis via Moran's $I$ corroborated this finding. The CNN representations exhibited artificially suppressed spatial consistency, whereas SAE dimensional expansion monotonically restored Moran's $I$ scores (Fig. 5b).

Consequently, this geometric contamination directly hindered downstream functional utility. Ridge regression to predict cell death rates revealed that the CNN representation's predictive capacity was artificially constrained; the SAE ($d = 8192, \lambda = 800$) consistently unlocked higher performance across all seeds and mutations (e.g., \emph{LRRK2} predictive $R^2$ improving from 0.288 to 0.345; Fig. 5(c)). UMAP projections visually confirmed this recovered structural integrity as a smooth functional gradient (Fig. 5d).

Collectively, these empirical findings validate our theoretical predictions: conventional dimensional bottlenecks inherently inflict severe geometric contamination, and unpacking this superposition is strictly necessary to faithfully capture the intrinsic semantic geometry of the data.

\vspace{-10pt} 
\section{Adaptation scRNA-seq Methodologies to Evaluate Representations}
\vspace{-4pt} 
\begin{figure}[t]
  \vspace{-20pt}
  \centering
  \nopagebreak
  \includegraphics[width=1.0\linewidth]{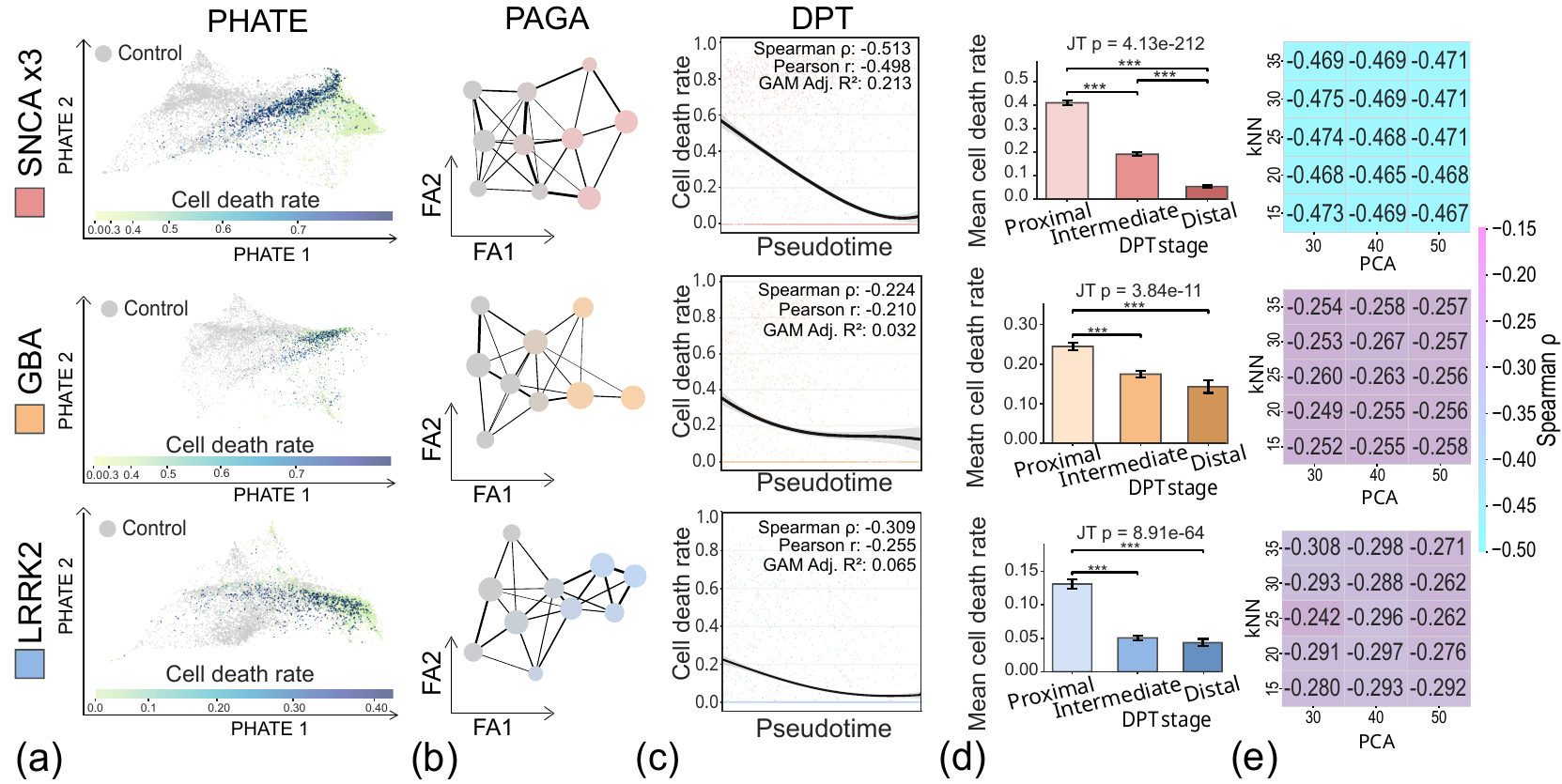}
  \captionof{figure}{Adaptation of single-cell manifold algorithms to SAE representations. All panels display pairwise comparisons between healthy controls (gray) and PD mutations (\emph{SNCA $\times 3$} $\times 3$: pink, GBA: orange, LRRK2: blue). \textbf{(a)} PHATE embeddings revealing continuous structure. \textbf{(b)} PAGA graphs with pie chart colors indicating the ratio of control to mutation cells in each node. \textbf{(c)} Scatter plots and GAM fits of cell death rates of individual images against geodesic Diffusion Pseudotime (DPT). \textbf{(d)} Average cell death rates across DPT-binned tertiles. \textbf{(e)} Robustness of the DPT-cell death correlation across varying hyperparameters}
  \vspace{-5pt}
  \label{PHATE,PAGA,DPT}
\end{figure}
By effectively resolving geometric contamination and mitigating cell density variation through IN as well as $L_2$ normalization, the SAE latent space functionally parallels the high-resolution topology of single-cell transcriptomic profiles. Intriguingly, the $L_0$/$L_1$ regularization imposed on SAEs mathematically mirrors the evolutionary constraints of energy-efficient, sparse molecular expression, yielding similarly skewed, zero-inflated activation distributions. Leveraging this structural similarity, we directly adapted scRNA-seq analytical frameworks to the image domain.

\enlargethispage{1\baselineskip}

Following standard scRNA-seq preprocessing (log-normalization and standardization), PHATE revealed continuous structures (Fig. 6(a)), while PAGA mapped the global connectivity between control and mutation states (Fig. 6(b)). Biologically, cells undergoing cell death converge toward a uniform, generic phenotypic state \cite{oflanaganDissociationSolidTumor2019, kopylovaConvergentTranscriptomicSignature2026}. Consequently, we hypothesized that highly toxic cells collapse toward the intersecting boundary of the control and mutation manifolds. That is, mutation cells closer to the healthy medoid exhibit higher cell death rates, whereas more distant cells remain viable. To quantify this trajectory, we calculated geodesic distances for mutation cells via diffusion pseudotime (DPT) rooted at the control medoid.

As anticipated, Generalized Additive Model (GAM) fitting demonstrated that cell death rates significantly decreased as cells diverged further from the control medoid across all mutations ($p < 0.001$; Fig. 6(c), (d)). This negative correlation was robustly maintained across diverse hyperparameters (Fig. 6(e)). Notably, \emph{SNCA $\times 3$} neurons exhibited the steepest trajectory ($\rho = -0.513$), aligning with prior literature that establishes $\alpha$-synuclein pathology as the primary driver of progressive cellular toxicity \cite{ludtmannAsynucleinOligomersInteract2018}. (Appendix F.1 for trajectory reliability) 

Taken together, these findings suggest that SAE-decontaminated image vectors can be reliably deployed as transcriptomic-like profiles, successfully bridging non-linear scRNA-seq analytical frameworks with high-content biological imaging. Crucially, translating high-content cellular images into single-cell state biological vectors circumvents chronic scRNA-seq artifacts—such as stochastic dropouts \cite{stegleComputationalAnalyticalChallenges2015, hicksMissingDataTechnical2018}, doublets \cite{wolockScrubletComputationalIdentification2019}, and cell-type biases \cite{vandenbrinkSinglecellSequencingReveals2017} while offering unprecedented cost and time efficiency.

\section{Cross-Modal Integration of Image and scRNA-seq Data via GW-map}

Because cellular images and single-cell transcriptomes are multimodal projections of the identical underlying biological state, their intrinsic geometric structures should be preserved irrespective of data modality. Given that SAE disentanglement resolves geometric contamination, we applied GW-map to couple these image representations with authentic scRNA-seq data. This coupling successfully bridged the modalities, achieving label transfer accuracies exceeding random chance (50\%) for both strictly one-to-one and probabilistic barycentric matchings (barycentric accuracy $>70\%$; Fig. 8(a)).

\begin{figure}[!t]
\centering 
\nopagebreak
    \vspace{-20pt}
    \includegraphics[width=1.0\linewidth]{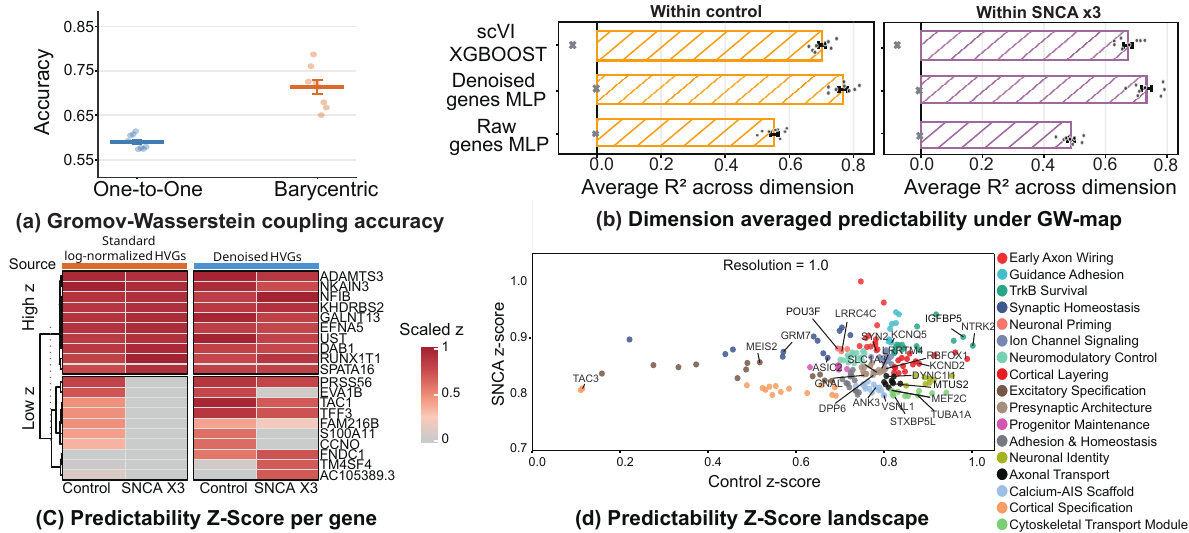} 
    \captionof{figure}{Cross-modal alignment via GW-map. \textbf{(a)} Label transfer accuracy of unconstrained GW coupling. One-to-one accuracy measures exact maximum-probability matches; barycentric accuracy evaluates probability-weighted matches. \textbf{(b)} Predictive performance ($R^2$) of XGBoost and MLP predicting scVI latent dimensions, standard log-transformed genes expressions, and scVI-denoised expressions directly from image representations. \textbf{(c)} Gene-specific predictability ranked by $Z$-score ($Z_{SNCA\times3}$ of standard log-normalized expression), highlighting top/bottom 10 genes. \textbf{(d)} 2D predictability landscape of the top 200 $Z_{SNCA\times3}$ genes reconstructing biological functional modules.}
    
    \vspace{-3pt} 
    \label{Fig5_GW_map_accuracy} 
\end{figure}

\enlargethispage{1\baselineskip}
To evaluate this alignment while accounting for genotype-specific confounding factors, we constrained couple vectors exclusively within identical biological classes (intra-class coupling). We then trained simple non-linear regressors (XGBoost and Multilayer Perceptrons) to predict transcriptomic profiles directly from the SAE image representations. Both models achieved remarkably high predictive fidelity ($R^2$) across scVI latent dimensions, standard log-normalized expressions, and scVI-denoised expressions, significantly outperforming permutation null baselines (mean permutation $R^2 < 0.1$). Notably, MLP-based prediction of scVI-denoised genes yielded $R^2$ values exceeding 0.7 across both classes (Fig. 8(b)).

If this alignment is biologically faithful, the capacity to predict a specific gene's expression from an image should proportionally reflect the gene's pathological relevance. To test this hypothesis while preventing variance-induced inflation of $R^2$ values, we quantified gene-specific predictability using $Z$-scores derived from the permutation null distributions. Ranked by predictability ($Z_{SNCA\times3}$ of standard log-normalized expression), the top genes included members of the 'axon guidance' cluster (e.g., \emph{EFNA5}, \emph{DAB1}) intimately associated with $\alpha$-synuclein pathological progression \cite{dijkstraEvidenceImmuneResponse2015.6.18} (Fig. 8(c)). Conversely, the least predictable genes (e.g., \emph{S100A11}, \emph{PRSS56}, \emph{FNDC1}, \emph{TM4SF4}) were non-specific markers of tissue damage or inflammation lacking distinct morphological signatures.

Furthermore, because genes operate within co-regulated networks, plotting genes in a ($Z_{Control}$, $Z_{SNCA\times3}$) predictability coordinate space would reveal a pathway-specific topography. Projecting the top 200 $Z_{SNCA\times3}$ genes onto a 2D landscape (Fig. 8(d)) unveiled an exceptional alignment with \emph{SNCA $\times 3$} mutation-associated neuronal architecture. Unsupervised Leiden clustering within this predictability space precisely reconstructed functional biological modules, notably isolating 'Presynaptic Architecture' and the adjacent 'Calcium-AIS Scaffold' pathways.

Taken together, these results demonstrate the high fidelity of GW-map and reveals that resolving geometric contamination via SAE allows imaging to integrate with authentic single-cell transcriptomics.

\section{Discussion}

Superposition is critical to address for two fundamental reasons. First, from an interpretability perspective, disentangling superposition allows for the direct interpretation of internal network representations, bypassing the manifold-attribution paradox. Moreover, while superposed networks exhibit continuous rotational symmetry, permitting arbitrary coarse isometric embeddings, compressed sensing theory demonstrates that explicit sparsity constraints break this continuous symmetry \cite{elhageToyModelsSuperposition2022b, donohoOptimallySparseRepresentation2003, donohoCompressedSensing2006}. This mathematically aligns the representational axes with objective biological bases, enabling data-aligned interpretation.

Second, while numerous studies on superposition are predominantly focused on interpretability, we revealed that superposition fundamentally corrupts the representational metric spaece. Given the importance of constructing representations that capture the underlying semantics, geometric contamination deserves further attention. By recovering this geometric fidelity, we successfully adapted scRNA-seq analytical frameworks directly to image domain. Given the structural similarity, scRNA-seq methods such as high-dimensional visualization, and gene co-expression networks can be directly adapted. Finally, we aligned SAE representations with authentic scRNA-seq data via GW-map with high biological plausibility. This establishes a powerful new paradigm for scalable spatial biology while corroborating that SAE representations faithfully reflect the intrinsic semantics of data.
\enlargethispage{1\baselineskip}
\paragraph{Limitations and Future Directions} First, while we demonstrated this framework using contrastive CNN representations, further investigation is required to validate our framework in other architectures, such as Vision Transformers (ViTs) \cite{dosovitskiyImageWorth16x162021}. Second, future studies should validate whether this geometric corruption and subsequent SAE-based recovery hold consistently across diverse data modalities and distinct clinical domains. Finally, while our GW-map framework provides a powerful \emph{de novo} cross-modal alignment, both training overcomplete SAE dictionaries and computing Gromov-Wasserstein optimal transport incur additional computational overhead.

\clearpage

\textbf{Acknowledgments}

This work was supported by the National Research Foundation of Korea,
funded by the Korean government (RS-2023-00266872 to M.L.C. and D.K.;
RS-2024-00343012 to M.L.C.; RS-2025-00521226 to D.K.). This work was
supported by the InnoCORE program of the Ministry of Science and ICT
(M.L.C.; N10260004). M.L.C. and D.K. acknowledge support from the KAIST
-- Formosa R\&D Center.

\textbf{Author Contributions}

J.P. and M.L.C. conceived and designed the study. J.P. developed the
software code, performed the experiments, and wrote the original draft
of the manuscript. M.L.C. supervised the study, wrote, and edited the
manuscript. S.K. performed the data analysis and generated the figures.
D.Y. conducted the cellular experiments. E.L., S.C., W.C., and I.C.
contributed to the data analysis. J.E. provided critical intellectual
input. D.K. contributed to the conceptualization of the study, provided
critical input, and edited the manuscript. S.G. provided the data
sources and edited the manuscript. All authors reviewed and approved the
final version of the manuscript.

\textbf{Conflicts of Interests}

The authors declare no conflicts of interest.

\textbf{Data Availability Statement}

The image dataset has been deposited in Zenodo, with a preview subset
available immediately, and the full dataset will be released upon
publication of the manuscript.

\bibliographystyle{unsrtnat}
\bibliography{overleaf_paper}
\clearpage

\appendix

\clearpage
\section{Cell Density Variation Normalization via GAP-L2 Scaling}
\subsection{Theoretical Analyis}

As established in previous studies \cite{gutTrajectoriesCellcycleProgression2015, heIntegratingSpatialGene2020}, variations in cell density can bias model representations. We denote the feature vector as $\mathbf{v} = (v_1, v_2, \dots, v_n) \in \mathbb{R}^n$, where each dimension $v_i$ is influenced by several factors: cell density ($d$), the biological state of cells ($b$), and the stochasticity of the training process. Since the weight matrix is normalized by its per-output-channel $L_2$ norm (see Appendix C. for detailed methods), the influence of training stochasticity is significantly mitigated. Consequently, $v_i$ can be expressed as a function of the remaining dominant factors:

\begin{equation}
    v_i \coloneqq f_i(d, b)
\end{equation}
\vspace{1em}

To understand the role of cell density, consider a scenario where $b$ is held constant while $d$ varies. It is plausible to think this scienario because the microscopic imaging process is independent of the state of the cell state. Given the local receptive fields of CNNs \cite{lecunGradientbasedLearningApplied1998, olahZoomIntroductionCircuits2020, burgertImageNettrainedCNNsAre2025a, brendelApproximatingCNNsBagoflocalFeatures2019a, dosovitskiyImageWorth16x162021, raghuVisionTransformersSee2021}, the feature extraction process can be approximated as a homogeneous function with respect to $d$ because GAP is linear. Specifically, we assume:

\begin{equation}
    f_i(nd, b) \approx n f_i(d, b) \quad \text{for any } n \in \mathbb{R}^+ \text{ and } (d, b) \in \mathcal{D}
\end{equation}
\vspace{1em}

This linear scaling property allows us to decompose the representation as $v_i \approx d \cdot \hat{f}_i(b)$. Under this assumption, the $L_2$ normalization of the feature vector $\mathbf{v}$ yields:

\vspace{1em}
\begin{equation}
    \frac{\mathbf{v}}{\|\mathbf{v}\|_2} \approx \frac{d \cdot \mathbf{\hat{f}}(b)}{\sqrt{\sum_{i=1}^n d^2 \hat{f}_i^2(b)}} = \frac{\mathbf{\hat{f}}(b)}{\sqrt{\sum_{i=1}^n \hat{f}_i^2(b)}}
\end{equation}
\vspace{1em}

As shown, the term $d$ is canceled out, rendering the final representation invariant to cell density. This ensures that the model captures the intrinsic biological state $b$ without being confounded by density variations.

\subsection{Emperical Validation}

As theoretically predicted, if $L_2$ normalization technically conditions out the influence of cell density variations, the representation will reflect the underlying semantics more accurately. To validate this, we trained ridge regression and XGBoost on CNN representations. For CNN which represetntaion is $L_2$ normalized before the projector, normalization substantially improved $R^2$ values across all mutation lines. (Fig. 9 (a))

Even in models trained without normalization (trained with three different random seeds), post hoc $L_2$ transformation consistently enhanced prediction accuracy ($R^2$), with statistically significant improvements observed for SNCA ×3 XGBoost, LRRK2 ridge regression, and LRRK2 XGBoost models (Fig. 8 (b)). Overall representational geometry was likwse preserved measured by CKA similarity between normalized and non-normalized models (Fig. 9).

\clearpage

\begin{figure}[t]
  \centering
  \nopagebreak
  \includegraphics[width=1.0\linewidth]{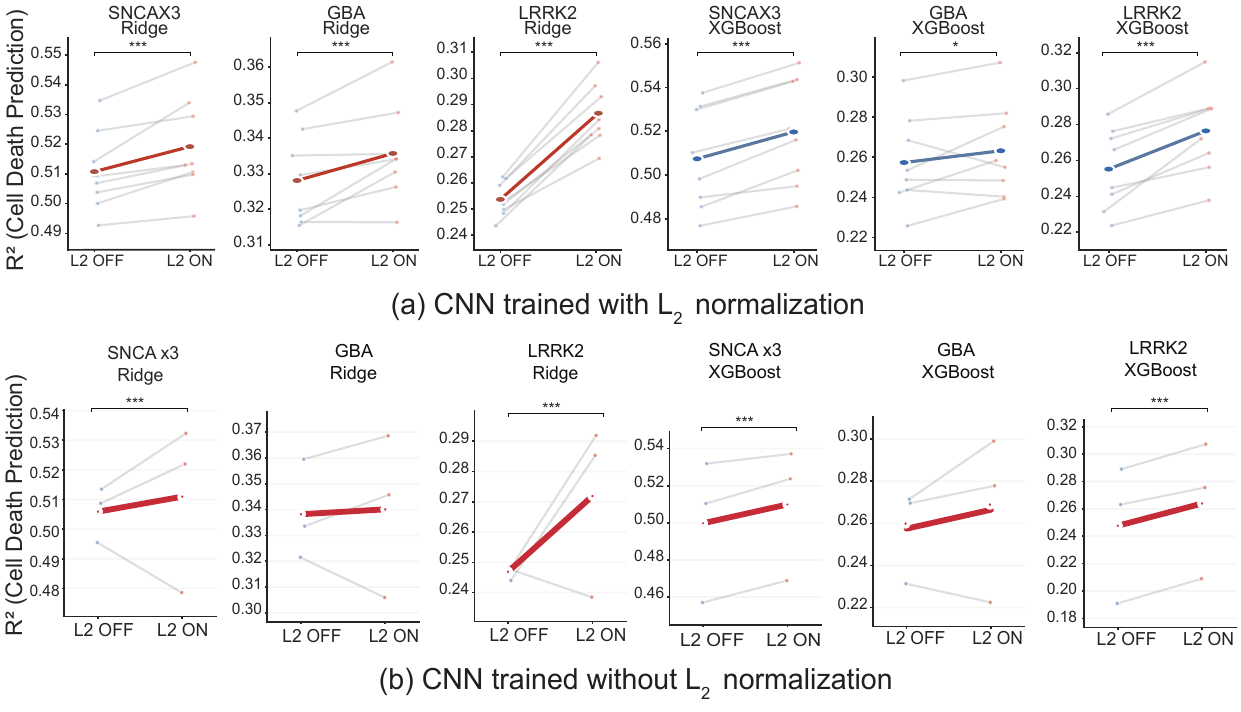}
     \captionof{figure}{\textbf{(a)} Effect of $L_2$ normalization on feature vectors derivend from CNN which feature vectors were $L_2$ normalized during training. showing It shows sensitivity to image cell-density variation, as quantified $R^2$ for cell death rate prediction using Ridge regression and XGBoost. \textbf{(b)} Effect of $L_2$ normalization on feature vectors derivend from CNN which feature vectors were not $L_2$ normalized during training. SNCA ×3 XGBoost, LRRK2 ridge regression, and LRRK2 XGBoost models were statistically significance.
    Statistical significance was assessed using a two-sided Wilcoxon signed-rank test, with each cross-validation fold treated as an independent observation (*p < 0.05, **p < 0.01, ***p < 0.001)}
    \label{L2 normalization effect}
    \vspace{-10pt}
\end{figure}

\begin{figure}[h]
\centering 
\nopagebreak
    \includegraphics[width=0.7\linewidth]{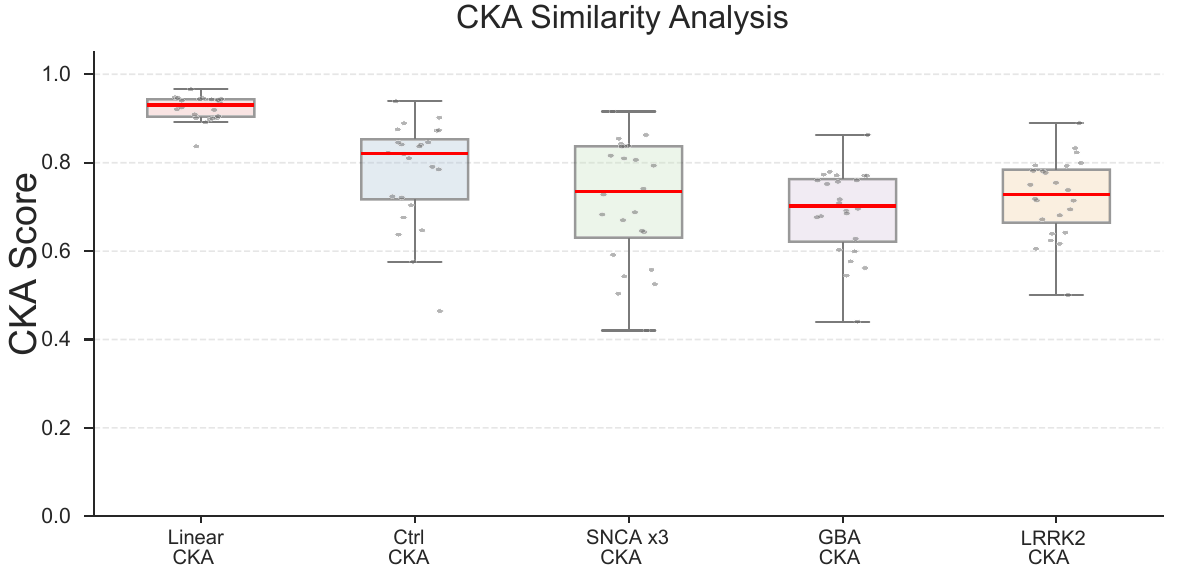} 
    \captionof{figure}{CKA analysis comparing models trained with and without L2 normalization on feature vectors. CKA analysis reveals that the model similarity remains considerably consistent regardless of whether GAP and L2 normalization are applied to the feature vectors before projector during training. This consistency suggests that L2 norm does not reflect the biological state. CKA was calculated 2500 images per class}
    \label{CKA_Boxplot_Organized_Points_norm_vs_noNorm} 
    \vspace{-10pt}
\end{figure}

Collectively, these results demonstrate that normalization effectively suppresses spurious density-dependent variation and enhances biologically meaningful latent representations. Consequently, all subsequent analyses were performed using $L_2$ normalized CNN feature vectors.

\section{Theoretical Analysis of Geometric Interference Due to Superposition}
\subsection{Strategic Superposition under dimensional bottleneck}

In modern neural architectures, it is plausible to assume that the number of concepts $n$ in real world typically exceeds the model's hidden dimension $d$ ($n >>d$). Recent research \cite{elhageToyModelsSuperposition2022b} shows that this dimensional bottleneck forces the model to employ superposition.

\noindent \textbf{Definitions:} 
Let $v_i \in \mathbb{R}^n$ be input features. The encoding process into the hidden space $\mathbb{R}^d$ is given by $\sigma(Wv_i + b)$, where $W \in \mathbb{R}^{d \times n}$ ($d < n$) and $\sigma(\cdot)$ denotes the ReLU activation function. By introducing a diagonal matrix $D_i$ representing the activation state of ReLU for $v_i$, the resulting representation $h_i \in \mathbb{R}^d$ is:
\begin{equation}
    h_i = \sigma(Wv_i + b) = D_i (Wv_i + b)
\end{equation}

\begin{equation}
    (D_i)_{kk} = 
    \begin{cases} 
    1 & \text{if } (Wv_i + b)_k > 0 \\
    0 & \text{otherwise}
    \end{cases}
\end{equation}

\noindent \textbf{Derivation:} Since $\text{rank}(W) \leq d < n$, the model cannot form an identity mapping for all $n$ input concepts. Following recent studies \cite{elhageToyModelsSuperposition2022b},  as ReLU and bias gating can cancel out trivial noise, the model opts for superposition rather than concepts discarding. Therefore, this rank deficiency ensures that any readout or distance calculation in $\mathbb{R}^d$ is fundamentally contaminated by the overlapping projections of non-orthogonal concepts.

The inner product between two representations $h_i, h_j \in \mathbb{R}^d$ inherits this structural interference:
\begin{equation}
    \langle h_i, h_j \rangle = v_i^T \underbrace{(W^T D_i D_j W)}_{\Phi_{ij}} v_j + \mathcal{C}_{ij}(W, D, b)
\end{equation}
where $\Phi_{ij} \in \mathbb{R}^{n \times n}$ is the effective projection matrix and $\mathcal{C}_{ij}$ aggregates bias-related noise terms ($b^T D_i D_j W v_j + v_i^T W^T D_i D_j b + b^T D_i D_j b$). 

Due to the bottleneck, $\text{rank}(\Phi_{ij}) \leq d < n$, forces $\Phi_{ij}$ into the form $I_n + E_{ij}$ where $E_{ij}$ $\neq 0$ represents the structural error matrix. Consequently, the inner product in the latent space is coupled with both rank-induced interference and bias noise:
\begin{equation} 
\langle h_i, h_j \rangle = \langle v_i, v_j \rangle + \underbrace{v_i^T E_{ij} v_j}_{\text{Rank-induced Interference}} + \underbrace{\mathcal{C}_{ij}(W, b, D)}_{\text{Bias Noise}} 
\end{equation}

The squared Euclidean distance between these latent representations $h_i, h_j \in \mathbb{R}^d$ is:
\begin{equation}
    \|h_i - h_j\|_2^2 = \|h_i\|_2^2 + \|h_j\|_2^2 - 2\langle h_i, h_j \rangle
\end{equation}
Consequently, this confirms that the metric itself, computed directly on the $d$-dimensional bottleneck, is a function of the superposition noise $E_{ij}$.

\subsection{\texorpdfstring{$L_p$}{L\_p} Norm Generalization via Local Norm Equivalence}
As discussed in section 3.4.1, superposition contaminates the Euclidean distances. The geometric contamination induced by the dimensional bottleneck is a metric-invariant flaw that propagates across the representation space. Within the local tangent space $T_p\mathcal{M}$ at any given semantic context $p$, the neighborhood behaves as a local linear vector space. Consequently, by the equivalence of norms ($C_1 \|h\|_2 \leq \|h\|_p \leq C_2 \|h\|_2$), the local geometric packing stress governed by the Generalized Welch Bound inherently permeates all local $L_p$ spaces ($p \geq 1$). Substituting the coupled local inner product expansion (Eq. 10) into the $L_2$ distance metric yields:
\begin{equation}
    \|h_i - h_j\|_2 = \sqrt{\|h_i\|_2^2 + \|h_j\|_2^2 - 2(S_{ij} + G_{ij})}
\end{equation}

where $G_{ij} = \sum_{k \neq l} v_{i,k} v_{j,l} (\mathbf{E}_p)_{k,l}(d)$ represents the coupled geometric contamination and $S_{ij} =\sum_{k=1}^m v_{i,k} v_{j,k}(\mathbf{G}_p)_{k,k} + \sum_{k \neq l} v_{i,k} v_{j,l}$ represents the pure local semantic Component as discussed in section 3.4.1. The dual nature of this local packing error $\mathbf{E}_p(d)$ establishes two fundamental boundaries that corrupt the metric space.

\paragraph{The Local Resolution Ceiling (Upper Bound on Separation)} 
For semantic instances composed of locally disjoint ground-truth concepts within $T_p\mathcal{M}$ (i.e., $S_{ij} = 0$), the latent distance should ideally reach its unconstrained orthogonal maximum ($h_i \perp h_j$). However, because the Generalized Welch Bound guarantees a strict positive distortion energy ($\|\mathbf{E}_p(d)\|_F^2 > 0$) due to the bottleneck $m > d$, severe structural collisions ($(\mathbf{E}_p)_{k,l}(d) > 0$) become mathematically unavoidable. For instances activating these colliding concept pairs, the positive local interference $G_{ij} > 0$ forces an artificial proximity, capping the achievable separation even under other $L_p$ metrics:
\begin{equation}
    \|h_i - h_j\|_p \leq C_2 \sqrt{\|h_i\|_2^2 + \|h_j\|_2^2 - 2 \cdot \text{Collision}(\mathbf{E}_p)} < C_2 \sqrt{\|h_i\|_2^2 + \|h_j\|_2^2}
\end{equation}
This ceiling confirms that the compressed representation space fails to provide sufficient white space for true semantic orthogonality, artificially pulling unrelated instances together and truncating contrastive resolution.

\paragraph{The Geometric Noise Floor (Lower Bound on Proximity)} 
Conversely, minimizing contrastive loss under the dimensional bottleneck forces the local dictionary $\mathbf{W}$ into a tightly packed, generalized ETF-like configuration, triggering ubiquitous background repulsion ($(\mathbf{E}_p)_{k,l}(d) < 0$) across non-correlated concepts to optimize the packing density. Because each instance $h_i$ is a linear combination of multiple distributed concepts, the probability that all cross-terms in $G_{ij}$ vanish simultaneously is asymptotically zero. This persistent negative contamination ($G_{ij} < 0$) creates an inescapable geometric noise floor that artificially inflates proximity, preventing identical semantic states from achieving perfect metric convergence:
\begin{equation}
    \|h_i - h_j\|_p \geq C_1 \sqrt{\|h_i\|_2^2 + \|h_j\|_2^2 - 2(S_{ij} - |\text{Repulsion}|)} > 0
\end{equation}
Consequently, the local $L_p$ metric space remains permanently distorted, failing to preserve the true unconstrained isometry of the underlying semantic manifold.

\subsection{Conclusion: Metric Entrapment}
Consequently, any $L_p$ measurement is trapped between a geometric noise floor that obscures fine-grained similarities and a resolution ceiling that prevents true semantic separation. The latent distance does not measure pure semantic proximity; rather, it reflects the equilibrium between structural collisions and heterogeneous repulsive packing. This metric entrapment mathematically necessitates the use of sparse autoencoders (SAEs) to expand the dimensionality to $M \geq m$, thereby ecoupling superposed features ($w_k^T w_l \to 0$ asymptotically), nullifying $G_{ij}$, and fully restoring the integrity of semantic distances.

\section{Thoretical Analysis of Metric Restoration SAE Sparsity}

\subsection{Mechanism of Noise Nullification:}
SAE resolve the rank deficiency by projecting bottlenecked representations into an expanded latent space $\mathbb{R}^M$ ($M \gg d$). This expansion, coupled with extreme sparsity, restores orthogonality by providing sufficient white space to decouple superposed concepts. The interference terms derived in Section 1, governed by $D_i D_j$, are theoretically nullified: letting $p = K/M$ be the activation probability ($K \ll M$), the expected overlap evaluates to:
\begin{equation}
    \sum_{k=1}^M \mathbb{E}[(D_i)_{kk}(D_j)_{kk}] = M \cdot p^2 = \frac{K^2}{M}
\end{equation}
As $M \to \infty$ and $p \to 0$, the overlap $K^2/M$ vanishes. This ensures $D_i D_j \approx \mathbf{0}$, which effectively orthogonalizes the concept dictionary ($w_k^T w_l \to 0$). Consequently, the Geometric Contamination ($G_{ij}$) from Eq. (10) is nullified, eliminating both rank-induced interference and structural noise.

\subsection{Geometric Flattening and \texorpdfstring{$L_p$}{L\_p} Integrity:}

Sparsity induces geometric flattening by isolating the active latent concept set within linear, low-dimensional subspaces. In this decompressed space $\mathbb{R}^M$
, the expanded dimension $M$ allows the $m$ ground-truth concepts superposed within the bottleneck ($d<m<M$) to transition from a crushed Softmax Code-like state to near-orthogonality. This expansion breaks the Metric Entrapment shown in Appendix B; by satisfying M>m, the resolution ceiling is lifted, and the geometric noise floor is cleared.

Letting $z_i \in \mathbb{R}^M$ be the sparse representation, any $L_p$ metric ($p \geq 1$) now faithfully reflects the pure semantic distance of the ground-truth concepts $v$:
\begin{equation}
    \|z_i - z_j\|_p \approx \left( \sum_{k \in \text{active}} |v_{i,k} - v_{j,k}|^p \right)^{1/p}
\end{equation}
This structural restoration ensures that latent distances measure the singular presence of semantic units, rather than the equilibrium of forced inter-feature competition. 

\subsection{Conclusion} 
In the dimensionality bottleneck ($d < m$) regime, generalized welch bound ensure a non-zero interference Consequently, latent distances in $\mathbb{R}^d$ remain trapped between a geometric noise floor and a resolution ceiling, failing to reliably reflect pure semantic proximity. This implies that such a vulnerability could be inherent to many representation learning paradigms, extending to modern architectures like CNNs, ViTs and transformers. This intrinsic metric corruption indicates that optimization alone cannot restore feature independence; instead, a transition to high-dimensional sparse representations via SAEs is required to nullify interference. Advanced architectures like Gated SAE further optimize this restoration by decoupling feature detection from magnitude estimation, thereby eliminating $L_1$ shrinkage bias and non-linear shifts. Consequently, SAEs restore the metric integrity and local linearity required for reliable manifold learning, clustering, and graph-based semantic analysis.

\clearpage

\section{Biological Plausiblity of GW-Map}
\begin{figure}[t]
  \includegraphics[width=1.0\linewidth]{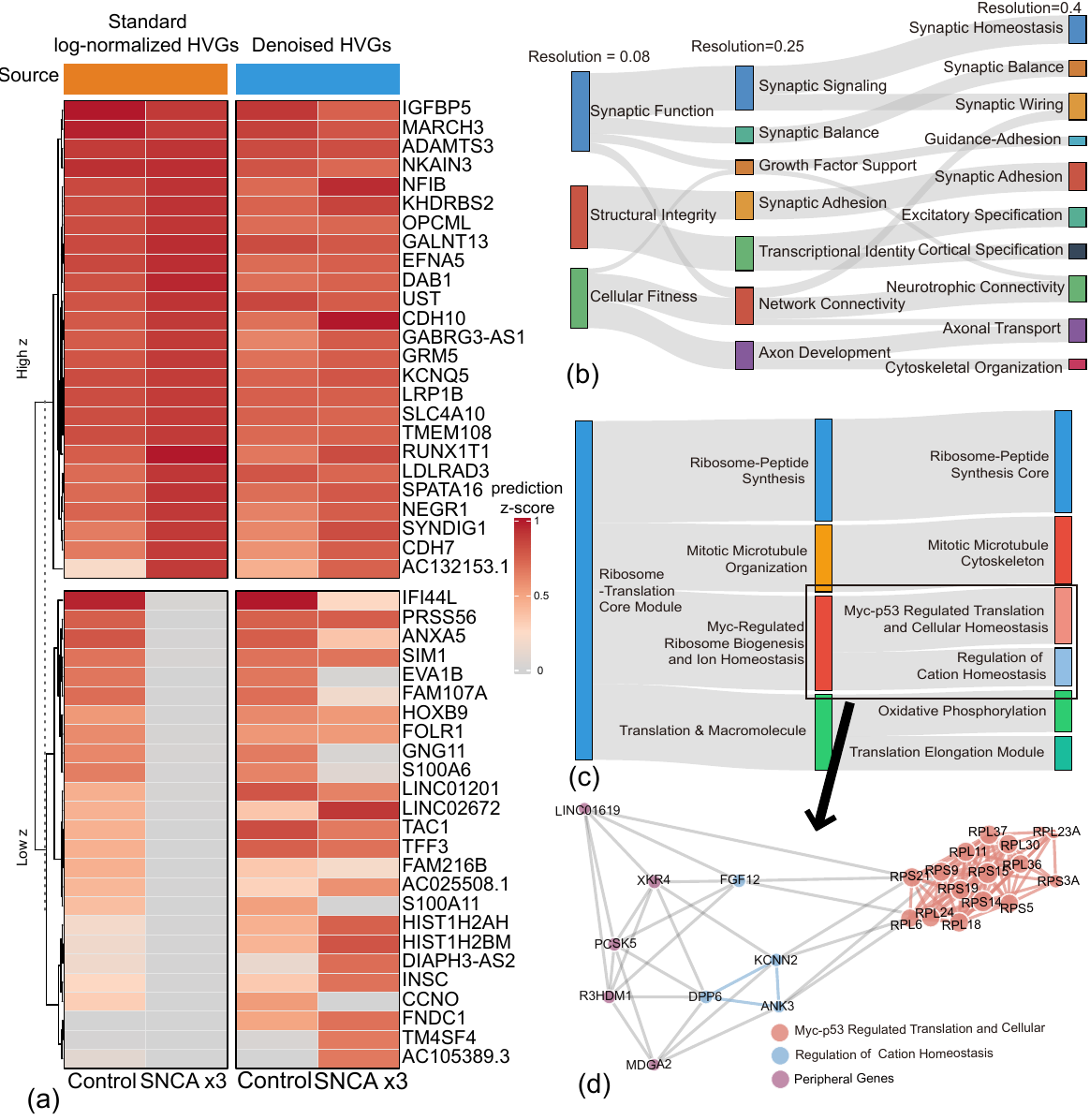}
     \captionof{figure}{\textbf{(a)}. Expression heatmap showing Z-scores (standard log-transformed and denoised) for $SNCA \times 3$ and Control intra-class coupling across the top and bottom 25 genes, which were selected based on $SNCA \times 3$ intra-class coupling Z-scores of standard log-normalized HVGs (computed against permutation-based null distributions). \textbf{(b)}. Sankey diagram tracking the hierarchical emergence of functional gene modules among the top 200 genes across increasing Leiden resolutions (0.08, 0.25, 0.4). The genes was selected based on $SNCA \times 3$ intra-class coupling Z-scores of standard log-normalized HVGs. Subsequent analyses were also conducted using these selected genes \textbf{(c)}. Global transcriptional module discovery. All valid genes were Leiden-clustered (resolutions=0.002, 0.2, 0.3) and functionally annotated via over-representation analysis (ORA; GO, KEGG, MSigDB Hallmark; FDR < 0.05). \textbf{(d)} Force-directed gene interaction network (Fruchterman-Reingold) resolving the hierarchical bifurcation of a higher-order transcriptional module. Nodes are colored by their fine-grained sub-module identities derived from top-ranked significant ORA hits. Node size is proportional to graph degree. Genes lacking significant enrichment for either sub-module are depicted as peripheral nodes. <Note> For (b--d), Graph adjacency matrices constructed via k-NN on dimension-wise rank-normalized coordinates bounded to [0,1] to enhance k-NN stability. (k=5 for (b); k=4 for (c), (d)). Genes exhibiting >99.5\% sparsity (zero counts) were strictly excluded. Comprehensive cluster gene lists are provided in the https://github.com/jijihihi/Bio\_superposition}
    \label{GW-map_biological_faithfulness}
\end{figure}

We further conducted subsequent analysis to verify the biological faithfulness of coupling. Previous studies \cite{heIntegratingSpatialGene2020, xieSpatiallyResolvedGene2023, xueInferringSinglecellResolution2025, wayMorphologyGeneExpression2022} suggest that deep learning-derived phenotypical representations encode underlying molecular information. Following the same procedure described in Section 6 (Figure 7), we selected the top/botton 25 genes based on $SNCA \times 3$ intra-class coupling Z-scores of standard log-normalized HVGs (Fig. 10 (a)). This reveals that genes with high $Z_{SNCA \times 3}$ scores include ‘axon guidance’ cluster (e.g., EFNA5, DAB1, CDH10, CDH7) \cite{dijkstraEvidenceImmuneResponse2015.6.18} as well as the ‘synaptic transmission and ion channel activity’ \cite{schirinziEarlySynapticDysfunction2016} (e.g., GRM5, KCNQ5, NKAIN3). These pathways are associated with $\alpha$synuclein pathological progression. Conversely, genes with $Z_{SNCA \times 3}$ approaching near zero (e.g., S100A6, S100A11, PRSS56, FNDC1, TM4SF4) are generally non-specific markers of tissue damage or inflammation with lower specific pathological relevance.

Based on the coordinates of genes on the $Z_{control}$ and $Z_{SNCA \times 3}$ in the section 6 Figure 7 (d), increasing the Leiden clustering resolution (from res=0.08 to 1.0) induced a hierarchical refinement, transitioning from a broad 'Synaptic Function’ ontology \cite{schirinziEarlySynapticDysfunction2016} into specialized regulatory sub-modules (Fig. 10 (b)). At the finest granularity, this approach unveiled specific pathological vulnerabilities: the segregation of 'TrkB Survival’ \cite{kangTrkBNeurotrophicActivities2017} and ‘Axonal Transport’ \cite{chuAlterationsAxonalTransport2012} anchored by key effectors such as DYNC1I1 and MUTS2 and the isolation of the ANK3-centric 'Calcium-AIS Scaffold' \cite{leterrierAxonInitialSegment2018}, confirming that our image-based prediction scores recognize and align distinct biological states in 2D space.

To validate these findings across the entire transcriptome, we performed Gene Set Over-representation Analysis (ORA) across increasing resolutions (Fig. 10 (c)). GW-map precisely deconstructed the 'Ribosome-Translation Core' into its constituent regulatory axes. Notably, increasing resolution from res=0.2 to 0.3 induced a systematic bifurcation of 'Myc-Regulated Ribosome Biogenesis’ \cite{morcelleOncogenicMYCInduces2019} into a specialized 'Myc-p53 Regulated Translation' axis and a discrete 'Cation Homeostasis’ program. This specification directly aligns with the established coupling between nucleolar stress and p53-mediated checkpoints \cite{lohrumRegulationHDM2Activity2003}, while effectively isolating the concurrent collapse in ionic buffering and Oxidative Phosphorylation \cite{vosMitochondrialComplexDeficiency2022, pokotyloMetabolicDysregulationParkinsons2025}. Furthermore, force-directed layouts confirmed that functionally synergetic genes form dense, autonomous interaction hubs such as RPL/RPS77,80, KCNN273, and ANK378 (Fig. 10 (d)). This granular alignment with canonical pathways confirms the robust fidelity of GW-map in coupling the image and scRNA-seq data. 

Collectively, these findings demonstrate that CNN-SAE captures essential genetic information from cellular images. Furthermore, the autonomous reconstruction of the hierarchical architecture in SNCA ×3 mutation-associated neuronal pathology confirms the high fidelity of GW-map in coupling neuronal images with organoid transcriptomic data, thereby validating the biological integrity of the GW-coupling framework and supporting the notion that disease-specific information is intrinsically encoded across heterogeneous data modalities.

\clearpage

\section{Detailed Method}

\subsection{Data Preparation and Cell Death Quantification}
\paragraph{Generation of Human iPSC-derived Cortical Neurons and High-content Image Acquisition}

Figure 11 illustrates the data acquisition process. Human iPSC-derived cortical neurons from three healthy controls and Parkinson’s disease-associated mutation lines are used in this study. Cell lines used are listed in Supplementary Table 1. hiPSCs were maintained on Geltrex-coated plates in E8 or mTeSR. Cortical differentiation was induced through dual SMAD inhibition and neuronal cultures were maintained as previously described by Choi et al\cite{dsaPredictionMechanisticSubtypes2023}. All cell lines were routinely tested for mycoplasma contamination.

Cortical neuronal identity was verified by immunocytochemistry using antibodies against MAP2 (Abcam, \# ab32454) and SATB2 (Santa Cruz Biotechnology, \# SC-81376). Immunostaining was performed following previously described procedures.

Live-cell imaging was carried out as previously described by Choi et al. Cortical neurons were plated onto 96-well plates at 20,000-40,000 cells per well and maintained until imaging. Before imaging, cells were stained with 10$\mu$M Hoechst 33342, 25nM TMRM, 250nM LysoTracker Deep, and 500nM SYTOX green. Images were acquired using the Opera Phenix High-Content Screening System (PerkinElmer).

\begin{figure}[h!]
  \includegraphics[width=1.0\linewidth]{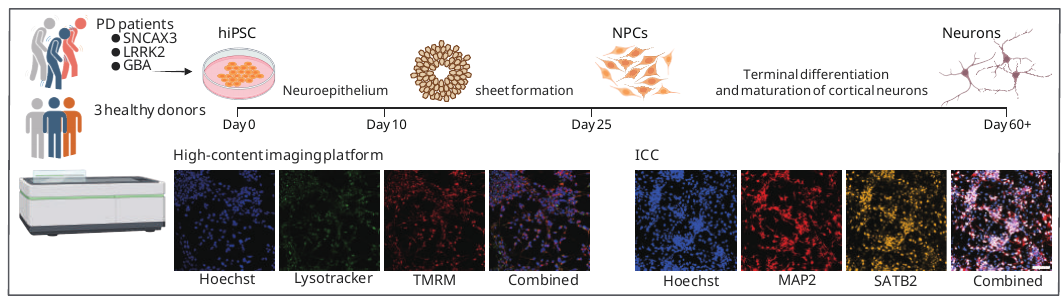}
     \captionof{figure}{Illustration of data acquisition. Cortical neurons derived from Parkinson’s disease patients were cultured, and data were acquired using a high-content imaging platform}
   \label{Data-collect}
\end{figure}

\paragraph{Image Preprocessing}
To fully utilize the 16-bit dynamic range while ensuring numerical stability during training, raw pixel intensities were divided by the maximum theoretical value (65535) to map them into the [0, 1] range. Following this scale transformation, each original $1024 \times 1024$ image was systematically partitioned into an $8 \times 8$ grid, generating non-overlapping $128 \times 128$ cropped patches. These cropped patches served as the localized input images for the downstream framework. To account for microscopy artifacts, such as variations in focus, and to normalize variations in cell density across different patches, we applied instance normalization to each individual input patch.

However, in cropped patches lacking specific organelles (e.g., the absence of nuclei in the blue channel due to uneven cell distribution), the pixel standard deviation within that channel can become exceptionally low. Applying standard instance normalization in these scenarios leads to the disproportionate amplification of background noise. To prevent this artifact, we implemented a clipped standard deviation approach, termed safenormalization: if the standard deviation of a given channel falls below an empirical threshold of 0.01, instance normalization is bypassed. This threshold was determined by systematically comparing patches with and without target organelles, thereby ensuring an optimal balance between signal preservation and noise suppression.

\paragraph{Quantification of Cell Death Rate}
To calculate the cell death rate, nuclear segmentation was initially performed on the $1024 \times 1024$ MIP images using StarDist (v0.9.1) with the pre-trained \texttt{'2D\_versatile\_fluo'} model. The segmentation utilized the default probability threshold (0.479) and a non-maximum suppression (NMS) threshold of 0.3. To ensure data integrity, segmented nuclei with an area smaller than 40 pixels were discarded as debris. 

Subsequently, the $1024 \times 1024$ images were cropped into $128 \times 128$ sub-images (patches). During this process, any $128 \times 128$ cropped images containing edge-touching nucleus relative to the original $1024 \times 1024$ boundaries were strictly excluded. This filtering step was necessary because nuclei located at the outermost boundaries of the original image might not be fully masked or segmented, which could introduce artifacts into the downstream cell death calculation. Furthermore, cropped images where nuclei occupied less than 25\% of the total pixels were removed to guarantee a sufficient cell density for a reliable estimation.

The Sytox Green threshold was then established via Multi-Otsu thresholding ($n=3$) based on the pixel intensities within the remaining nuclear masks. Among the two generated thresholds, the highest value was selected as the final criterion for Sytox Green-positive signals. Rather than relying on a binary cell-count metric (i.e., alive versus dead), which obscures the continuous nature of cell degeneration, the cell death rate was quantified by calculating the ratio of the total area of Sytox Green-positive pixels to the total nuclear pixel area within each valid patch. This area-based metric robustly captures the continuous, multi-stage progression of cellular apoptosis and necrosis.

\subsection{CNN Framework with Contrastive Learning }
\paragraph{Network Architecture}
To ensure spatial interpretability, we employ a CNN-based architecture rather than a Vision Transformer (ViT). This choice is motivated by the CNN’s local receptive field , which allows for a direct mapping between feature activations and specific image regions. We designed a custom CNN architecture to maintain the feature maps at a high spatial resolution of $64 \times 64$ ($H \times W$) for better interpretability. This architectural choice ensures that the upsampled activation maps via bilinear interpolation provide fine grained and precise localization of the identified features. The model consists of a stem layer followed by four stages and a refine stage. The stem layer utilizes a convolutional layer with a kernel size of 3 and a stride of 2, downsampling the 128×128 input to 64×64. The core architecture includes 9 residual blocks grouped into four stages (2, 2, 2, 3 blocks, respectively), with each block comprising two convolutional layers with skip connections to facilitate gradient flow. The network starts with 64 channels in the stem layer and progressively increases the feature depth to 128, 256, and 512 channels across stages 2 through 4, while stage 5 maintains 512 channels. A final refine stage containing one additional residual block was appended to further process the high-level semantic features. ReLU activation was used throughout the network. The output feature maps from the refine stage were global average pooled(GAP) and then selectively L2-normalized to adjust the cell density variation before being passed to the projector(Supplementary Note 1). The projector consists of two linear layers with a hidden dimension of 2048, mapping the 512-dimensional input to a 512-dimensional embedding space, which was then L2-normalized for the contrastive loss computation.

\paragraph{Contrastive Learning Framework via Moco}
The model was trained using Supervised Contrastive Learning (SupCon) combined with Momentum Contrast (MoCo). For each training step, two random rotated augmented views (v1, v2) were generated from each image. The query encoder ($model_q$) processed v1 to produce query representations, while the momentum encoder ($model_k$) processed v2 to produce key representations without gradient computation. The momentum encoder was updated via exponential moving average (EMA) of the query encoder parameters with a momentum coefficient of m = 0.995: 

The parameters are updated as follows:
\begin{equation}
\theta_k \leftarrow m \cdot \theta_k + (1 - m) \cdot \theta_q
\label{eq:update}
\end{equation}
where $m$ denotes the momentum coefficient. 

A large FIFO queue with a capacity of 65,536 entries stored historical key representations and their labels, enabling the contrastive loss to be computed against a large and diverse set of negatives beyond the current 512 mini-batch. At each step, the key pool consisted of the current batch keys concatenated with the queue. Subsequent analysis was conducted using only the query encoder after training.

\paragraph{CNN Training Strategy}
Since conventional augmentation techniques, such as Gaussian blurring, can compromise subtle textures or distort local morphological features of cellular images, we restricted our augmentation to random rotations to preserve fine-grained biological details. To mitigate the GAP fluctuation due to the stochastic weight updates during the training, we applied the L2 normalization every step.

For each Conv2d and Linear layer, the weight matrix is divided by its per-output-channel $L_2$ norm:
\begin{equation}
\mathbf{w} \leftarrow \mathbf{w} / \|\mathbf{w}\|_2
\end{equation}
This mitigate the GAP fluctuation due to the stochastic weight updates during the training. 

To prevent the model from learning batch-specific confounders such as plate effects or cell line imbalances, we employed a strictly balanced batch sampling strategy. Each mini-batch was constructed to contain an equal number of images from every combination of genotype class, cell line, and culture plate, ensuring that no single biological or technical covariate was over-represented within a batch.

We selected Stochastic Gradient Descent (SGD) over Adam to ensure compatibility with our per-iteration L2 normalization strategy. While Adam’s adaptive moment estimation adjusts individual weight scales independently, it can interfere with the constrained feature space imposed by the L2 norm. In contrast, SGD's consistent gradient updates allowed for more stable convergence. 

A linear warmup over 4 epochs was used, ramping the learning rate from 0 to the base learning rate of 0.1. After warmup, a cosine annealing schedule decayed the learning rate to the minimum value of 0.0. Gradient clipping with a maximum norm of 1.0 is applied.

The dataset was partitioned into training, validation, and testing sets with a ratio of 3:1:2, respectively. The model was trained using 8 seeds for L2-normalized projector inputs and 3 seeds for non-normalized projector inputs.

\paragraph{Loss Function}
Supcon loss was computed as:

\begin{equation}
L = - \frac{1}{|P(i)|} \sum_{p \in P(i)} \log \left[ \frac{\exp(q_i \cdot k_p / \tau)}{\sum_{j} \exp(q_i \cdot k_j / \tau)} \right]
\end{equation}

where $P(i)$ denotes the set of positive keys sharing the same genotype label as query $q_i$, $k_p$ represents a positive key, and $\tau = 0.07$ is the temperature parameter. Numerical stability was ensured by subtracting the maximum logit before exponentiation.

\paragraph{Linear Probing and Evaluation}
To evaluate the representation quality of the trained CNN framework, we conducted a linear probing analysis. Weight parameters in the CNN encoder were frozen after contrastive learning, ensuring a static weight configuration that was consistently applied to all subsequent analyses. A bias-free linear classifier ($\mathbb{R}^{512} \to \mathbb{R}^{4}$) was then attached to the $L_2$-normalized encoder outputs. 

The linear classifier was trained for 50 epochs on the training set without data augmentation, using SGD with a learning rate of 0.1, a momentum of 0.9, and no weight decay. At the end of each training epoch, the representation quality was monitored on the validation set. The checkpoint achieving the highest linear probe accuracy on the validation set was selected as the final model. The classification performance of this selected model was ultimately evaluated and reported using the held-out test set without augmentation.

\subsection{Representational Similarity Analysis via CKA}
To evaluate and compare the CNN model representations learned across different seeds, we utilize linear Centered Kernel Alignment (CKA) on the feature vectors of the $f_{\text{refine}}$ layer at both the class and global levels. The CKA similarity is formulated based on the Hilbert-Schmidt Independence Criterion (HSIC).

Let $K = XX^T$ and $L = YY^T$ denote the Gram matrices. The Linear CKA coefficient is computed as follows:
\begin{equation}
\text{CKA}(K, L) = \frac{\text{HSIC}(K, L)}{\sqrt{\text{HSIC}(K, K) \text{HSIC}(L, L)}}
\end{equation}
where the centered HSIC is defined as:
\begin{equation}
\text{HSIC}(K, L) = \frac{1}{(n-1)^2} \text{tr}(KHLH)
\end{equation}

In this expression, $H = I_n - \frac{1}{n}\mathbf{11}^T$ represents the centering matrix. The resulting CKA score ranges from 0 to 1, where a value of 1 indicates that the two models have captured identical representations. 

For this analysis, 2,500 images per class were used as input, with each image represented by a 512-dimensional feature vector. Given that the CKA analysis revealed a remarkably high architectural similarity between representations obtained with and without the $L_2$-normalized projector, we exclusively utilized the GAP $L_2$-normalized features for all subsequent analyses.

\subsection{Sparse Autoencoder}
\paragraph{Gated Sparse Autoencoder Architecture}
To resolve concept superposition, we utilize a gated sparse autoencoder(SAE) to decompose feature maps into monosemantic and interpretable components. This decomposition ensures that vector distances more faithfully reflect the underlying biological states. Unlike conventional sparse autoencoders that often suffer from shrinkage bias due to $L_1$ regularization, the gated sparse autoencoder architecture effectively decouples the activation magnitude through a gating mechanism. The dictionary size ($d_{sae}$) and its corresponding sparsity($\lambda$) were evaluated under various hyperparameter configurations and detailed evaluation metrics are shown in section 4.1

We integrated an SAE into a each standalone CNN encoder backbone. Each spatial location in the feature map is treated as an individual 512-dimensional vector, serving as the input token for point-wise SAE training. The input tokens are batch-centered(batch size N=64) and $L_2$-normalized. The same processing was applied to all subsequent analyses. The gating and magnitude encoders share the same weight matrix ($W_{gate} = W_{mag}$) while maintaining independent norms and biases, following the standard weight-tying configuration in SAE. The gating biases $b_{gate}$ are initialized with a small positive constant 0.1 to ensure every latent feature maps are sufficiently updated.
 
We constrain the columns of the decoder weight matrix ($W_{dec} \in \mathbb{R}^{d_{in} \times d_{sae}}$) to have exactly unit $L_2$ norm by renormalizing them after every update to prevent scaling ambiguity:
\begin{equation}
    \|\mathbf{w}_i\|_2 = 1, \quad \forall i \in \{1, \dots, d_{sae}\}
\end{equation}

Additionally, we project out the gradient components parallel to these rows before applying the optimizer update:
\begin{equation}
    \nabla_{\mathbf{w}_i}^{proj} \mathcal{L} = \nabla_{\mathbf{w}_i} \mathcal{L} - \left( \nabla_{\mathbf{w}_i} \mathcal{L} \cdot \mathbf{w}_i \right) \mathbf{w}_i
\end{equation}

The resulting gated activation $h$ is formulated as the element-wise product of the binary gate and the magnitude vector:
\begin{equation}
    h = \mathds{1}(g_{pre} > 0) \odot \text{ReLU}(m_{pre})
\end{equation}
where $\mathds{1}(\cdot)$ denotes the indicator function. The pre-activations are computed as:
\[
    g_{pre} = W_{gate}v + b_{gate}, \quad m_{pre} = W_{mag}v + b_{mag}
\]
Here, \textbf{$v \in \mathbb{R}^{d_{in}}$} represents the input activation vector and $W_{gate}, W_{mag} \in \mathbb{R}^{d_{in} \times d_{sae}}$ are the encoder weights for the gating and magnitude paths, respectively.

During backpropagation, the gating mechanism is implemented using a Straight-Through Estimator (STE); while the forward pass employs a binary gate based on a step function, the backward pass utilizes the sigmoid gradient as an approximation to ensure stable parameter updates.

\paragraph{Gated Sparse Autoencoder Loss Function and Train Strategy}

The SAE is trained by minimizing a loss function composed of reconstruction, sparsity, and auxiliary terms. The total loss $\mathcal{L}$ is defined as follows:
\begin{equation}
    \mathcal{L} = \mathcal{L}_{recon} + \lambda \mathcal{L}_{sparsity} + \alpha \mathcal{L}_{aux}
\end{equation}
where $\lambda$ and $\alpha$ are hyperparameters controlling the strength of the sparsity penalty and the auxiliary task, respectively. Sparsity warmup period was set to 100 steps. The auxiliary loss weight $\alpha$ is fixed at 0.03125.

The model was trained for 8 epochs with a total batch size of 262,144, which was processed in 4 distinct chunks to manage memory constraints. To prevent batch-level data bias and ensure a balanced feature distribution, tokens were randomly permutated prior to being input into the SAE. We employed an initial learning rate of $3 \times 10^{-4}$ with a 10\% learning rate warmup fraction, and a weight decay of $1 \times 10^{-4}$ regulated by a cosine scheduler.
The model from the final epoch is selected for downstream tasks. To prioritize regions identified as significant by the CNN, we selectively multiply the token L2 norm by the reconstruction loss for instances utilizing $\lambda$ = 3200

Reconstruction loss is computed as mean squared error(MSE).
\begin{equation}
    \mathcal{L}_{recon} = \|v - (W_{dec}h + b_{dec})\|_2^2
\end{equation}

Penalizing only the gating signal, the model learns sparse activation boundaries without suppressing the feature magnitudes:
\begin{equation}
    \mathcal{L}_{sparsity} = \sum_{i=1}^{d_{sae}} \text{ReLU}(g_{pre,i})
\end{equation}

To prevent the gating mechanism from becoming trapped in sub-optimal dead states, an auxiliary reconstruction loss is employed. Rather than simply using all magnitude information, we utilize a specialized computational path that trains only a subset of inactive features.
The auxiliary activations, $m_{aux}$, are formulated by masking out the currently active features and retaining only the top-$K$ instances (with $K=32$) based on their magnitude:
\begin{equation}
    m_{aux} = \text{Top}_K \Big( \text{ReLU}(m_{pre}) \odot \mathds{1}(g_{pre} \le 0) \Big)
\end{equation}
The auxiliary reconstruction loss is then computed using the stop-gradient operator ($\text{sg}(\cdot)$), which detaches the decoder weights and biases during backpropagation. This critical detachment prevents the auxiliary error gradients from inappropriately altering the main decoder weight distributions:
\begin{equation}
    \mathcal{L}_{aux} = \big\| v - \big( \text{sg}(W_{dec}) m_{aux} + \text{sg}(b_{dec}) \big) \big\|_2^2
\end{equation}

\paragraph{Evaluation of Gated Sparse Autoencoder}
For each input image, the pre-trained CNN activations are passed through the SAE to obtain a sparse feature map $h \in \mathbb{R}^{H \times W \times d_{sae}}$. To eliminate the randomness of batch centering and per-image artifacts, we perform centering at the token level within each individual image. The extracted feature representations were subsequently cached and utilized for all downstream evaluation analyses. We applied GAP and get $d_{sae}$-dimensional latent feature vector. Following the standard linear probing protocol, we train a single linear classification layer (a linear probe) on top of these pooled vectors while keeping the SAE and CNN weights frozen and test with a test images.

The feature activation frequency is tracked via an EMA (decay = 0.99). Linear probe is trained with SGD (momentum=0.9) optimizer and cross entroy loss. Because the L2 norm of CNN token is information indicating degree of focus the model assigns to specific features, we test two distinct representations: raw feature vector and its L2 norm-scaled counterpart. L2 norm-scaled vector was consistently employed in the following analytical steps yielding activation maps that preserve relative spatial magnitude. Finally, the global average-pooled SAE latent feature vectors were $L_2$-normalized. This step is conducted to correct for the inherent reduction in Euclidean magnitude that occurs when a singular representation is disentangled into multiple sparse variables (i.e., $|a| \approx |x|+|y|+|z|$, but $a^2 \ge x^2+y^2+z^2$), ensuring that highly fractionated concepts do not artificially lose their discriminative influence in downstream distance metrics.

We further quantified the Fraction of Variance Unexplained (FVU) and the mean inner product of the concept vector (column of SAE decoder). The FVU was computed by comparing the reconstructed outputs with the activations of the CNN to assess reconstruction fidelity.

\subsection{Predicting the Cell Death Rate using Feature Vector}

\paragraph{Feature Vector Extraction and Preprocessing}
We trained Ridge regression and XGBoost models to predict per-image cell death rates directly from CNN and SAE feature vectors. To eliminate genotype confounding effects, regression models were trained and evaluated independently within each genotype. Briefly, feature vectors were extracted from three layers of a single trained CNN encoder ($f_{stage5\_mid}$, $f_{stage5\_out}$, and $f_{refine\_out}$) and collapsed via GAP. For the SAE, pre-trained CNN  $f_{stage5\_out}$ activations for each input image were fed into the SAE to yield a sparse latent feature map, $h \in \mathbb{R}^{H \times W \times d_{\text{sae}}}$. We then applied GAP to obtain a $d_{\text{sae}}$-dimensional feature vector, while restoring the original $L_2$-norm magnitude within this latent representation. Consequently, all downstream representation analyses were performed using global average-pooled feature vectors.

The extracted features were then partitioned by genotype and matched with their respective cell death rate labels. To mitigate overfitting and ensure a fair comparison across disparate feature spaces, all feature dimensions were reduced to 250 via Principal Component Analysis (PCA) and standardized to zero mean and unit variance prior to downstream modeling.

\paragraph{Model Configurations} 
Ridge Regression was trained as a linear baseline. The regularization parameter was optimized via internal leave-one-out cross-validation, performing a grid search over 50 candidate values spaced evenly on a log scale from $10^{-3}$ to $10^{5}$.

XGBoost was trained to capture nonlinear relationships. we utilized a gradient-boosted tree ensemble (XGBRegressor). The hyperparameters were set as follows: number of estimators = 300, maximum tree depth = 5, learning rate = 0.04, row subsampling rate = 0.8, and column subsampling rate per tree = 0.8 (colsample\_bytree). The histogram-based tree method (tree\_method="hist") was used. Within each training fold, 15\% of the training samples were held out as an internal validation set for early stopping, with a patience of 30 rounds to prevent overfitting.

\paragraph{Validation and Robustness} 
We employed a repeated 5-fold cross-validation (CV) scheme to assess performance. All folds were stratified and shuffled with a fixed random seed. The coefficient of determination ($R^2$) was calculated for each fold as:
\begin{equation}
    R^2 = 1 - \frac{\sum_i (y_i - \hat{y}_i)^2}{\sum_i (y_i - \bar{y}_{\text{test}})^2}
\end{equation}
where $y_i, \hat{y}_i$, and $\bar{y}_{\text{test}}$ represent the observed labels, predicted values, and the mean of the test-fold labels, respectively. To ensure the reliability of our findings, the entire pipeline was repeated across 8 independent seeds for CNN and corresponding SAE on a fixed backhone. Additionally, a permutation test was conducted by shuffling apoptosis labels while maintaining the original feature structures and CV splits to establish a null distribution. Statistical significance between the CNN and SAE was determined by a two-sided Mann-Whitney U test and each seed treated as an independent observation.

\paragraph{Layer Comparision of CNN layers and {$L_2$} Nomralization Effect.} 
The efficacy of $L_2$ normalization (cell density correction) on CNN $f_{\text{stage5\_out}}$ representation was evaluated by quantifying the performance gain relative to raw features. Statistical significance for CNN layer comparisons,  normalization effects and comparision between CNN and SAE were assessed using a two-sided Wilcoxon signed-rank test.

\subsection{A Geometric Analysis of Vector spaces: Comparing CNN and SAE Representations}

\paragraph{Quantification of Representation Fidelity via $K$-NN Phenotypic Coherence and Moran's I}
If SAE representation more faithfully reflect the data intrinsic sturcture by disentangling the superposition, the variance of phenotypic labels within a $k$-nearest neighbor ($k$-NN) neighborhood is expected to be significantly lower than the global variance across the class. To quantify this local semantic consistency, we constructed $k$-NN graphs on both the CNN and SAE feature vectors across varying neighborhood sizes ($k \in \{5, 10, 15, 20, 25\}$), computed independently within each mutation class Phenotypic coherence was evaluated using the standard deviation ratio:
\begin{equation}
    \text{Ratio} = \frac{\sigma_{\text{local}}}{\sigma_{\text{global}}}
\end{equation}
where $\sigma_{\text{local}}$ is the standard deviation of the cell death rate within a given data point's $k$-NN neighborhood, and $\sigma_{\text{global}}$ denotes the baseline standard deviation across the each mutation class. A lower ratio indicates that the spatial clustering of the feature space closely aligns with semantic relatedness. To statistically compare the representational spaces, we computed the mean of the standard deviation ratios across all data points.

To complement the local dispersion metric, we evaluated the spatial autocorrelation of phenotypic labels using Global Moran’s I calculated per mutation class across the same $k$-NN defined neighborhoods. The statistic is defined as:
\begin{equation}
    I = \frac{n}{\sum_{i=1}^{n}\sum_{j=1}^{n} w_{ij}} \frac{\sum_{i=1}^{n}\sum_{j=1}^{n} w_{ij}(z_i - \bar{z})(z_j - \bar{z})}{\sum_{i=1}^{n} (z_i - \bar{z})^2}
\end{equation}
where $n$ is the total number of observations within the given class, $z_i$ represents the phenotypic label value of the $i$-th instance, $\bar{z}$ is the sample mean of that class, and $w_{ij}$ denotes the spatial weight matrix derived from the intra-class $k$-NN topology (where $w_{ij} = 1$ if $j$ is a $k$-nearest neighbor of $i$ ($i \neq j$), and $0$ otherwise). While the standard deviation ratio captures the local dispersion of labels within immediate neighborhoods, Global Moran’s I rigorously evaluates the overall spatial continuity across the entire intra-class manifold.

\paragraph{Assessing Information Capacity via Effective Rank}
To compare the information capacity of the CNN 3 layers and SAE feature spaces, we computed their Effective Rank ($e\text{Rank}$) under three settings: (1) raw high-dimensional features, (2) 250-dimensional PCA-reduced features, and (3) 250-dimensional PCA-reduced features with prior feature-wise standardization. PCA projection to a fixed dimension (250D) ensures a fair comparison by aligning the capacity ceiling across disparate architectures (e.g., CNN vs. overcomplete SAE). Additionally, the standardization step (scaling features to unit variance) was selectively applied to evaluate whether low-variance directions encode meaningful biological structures, preventing them from being dominated by a few high-variance components.

The $e\text{Rank}$ quantifies the effective dimensionality of a feature matrix based on the Shannon entropy of its variance distribution. Let $\lambda_i$ ($i = 1, \dots, m$) denote the eigenvalues of the covariance matrix sorted in descending order, where $m$ is the number of principal components. These eigenvalues are normalized to form a discrete probability distribution:
\begin{equation}
    p_i = \frac{\lambda_i}{\sum_{j=1}^{m} \lambda_j}
\end{equation}
The effective rank is defined as the exponential of this spectral entropy:
\begin{equation}
    e\text{Rank} = \exp\left( -\sum_{i=1}^{m} p_i \ln p_i \right)
\end{equation}
A higher $e\text{Rank}$ indicates a more uniform variance distribution across orthogonal dimensions, signaling higher information capacity. Conversely, an $e\text{Rank}$ close to 1 implies that the representational space has collapsed into a single dominant dimension.

\subsection{Adaptation of Single-cell seq Methods to the Representation Derived from image}
\paragraph{Visualization of Cell States via PHATE}
We employed Potential of Heat-diffusion for Affinity-based Trajectory Embedding (PHATE), which preserves both local and global data structure. SAE representations were treated as analogues of single-cell gene expression profiles: each image-level feature vector represents a point in the learned feature space, where local Euclidean distances reflect semantic similarity owing to the disentangled nature of the SAE representation.

To directly visualise the relationship between phenotypic progression and cell death, a separate PHATE embedding was constructed each control and mutations pair (Control-\emph{SNCA x3}, Control-\emph{GBA}, Control-\emph{LRRK2}). After log-transformation followed by z-score standardization (log-z-standardization), the filtered feature matrix was reduced to 50 dimensions via PCA. PHATE was then computed with $k = 5$ nearest neighbours, $\alpha$-decay $= 120$, Euclidean distance, and the diffusion time was set to 35.

\paragraph{Pseudotime Analysis and Partition-based Graph Abstraction}
We adopted diffusion pseudotime to SAE representations derived from images to compute trajectories of control-mutatnt pairs. To prevent circular analysis, the dataset was stratified into a feature-selection set (50\%) and an evaluation set (50\%); differential activation was computed on the former, and all downstream statistics were assessed on the latter.

Feature vectors were $L_2$-normalized and log-$z$-standardization. For each mutation, a Control-mutation cell pair was constructed. The cells were then projected into a diffusion map space with 15 components. A $k$-NN graph was constructed using Euclidean distance. These diffusion pseudotime (DPT) analyses were implemented using Scanpy's sc.tl.diffmap (with n\_comps=15) and sc.tl.dpt functions. The root cell was selected as the control cell closest to the control centroid in diffusion-map coordinates, serving as an approximation of the Fréchet centroid. Partition-based graph abstraction (PAGA) was employed to verify pathological continuity and connectivity between mutation and control clusters with resolution 0.65.

The monotonic association between DPT and apoptosis was assessed by Spearman's $\rho$ and Pearson's $r$. A generalized additive model (GAM; cubic splines) was fitted to capture nonlinear trends, and adjusted $R^2$ was computed to quantify the explained variance of the DPT--apoptosis relationship. To test the robustness of root cell selection, we performed a root perturbation analysis by randomly resampling $n = 10$ alternative root cells were selected from the neighborhood of the Control medoid and computing the resulting distribution of Spearman's~$\rho$. Cells were categorized into three DPT terciles based on the 5th–95th percentile range to exclude outliers. A Jonckheere–Terpstra (JT) test was applied to assess the significance of the ordered trend in apoptosis rates across these stages. Post-hoc pairwise comparisons (Proximal vs. Intermediate, Intermediate vs. Distal, and Proximal vs. Distal) were performed using Dunn’s test with Bonferroni correction for multiple comparisons.

\paragraph{Characterization of Feature Map Activation Dynamics along the Pseudotime Trajectory}

To visualize the continuous dynamics of the learned representations across DPT, feature vectors were first ordered by their DPT values and aggregated into 100 equal-sized pseudotime bins to compute stable mean feature activations. To emphasize the relative temporal dynamics, the binned activation profile of each feature was Z-score normalized and subsequently smoothed using a 1D Gaussian filter ($\sigma=1.5$). 

To identify modules of features exhibiting synchronized activation patterns, we constructed a 
$k$-NN ($k$ = 15) graph and applied the Leiden algorithm with a resolution of 1.0 on the smoothed temporal profiles. The resulting feature modules were then hierarchically ordered according to their peak activation time along the pseudotime trajectory, visualizing a sequential cascade of phenotypic transformations from the early (Control-like) to late (pathological) stages. 

Furthermore, to quantify model the continuous, non-linear activation trends of interpretable individual feature map activation along the DPT axis, we fitted GAM utilizing penalized cubic regression splines. To mitigate the potential leverage of extreme density tails, the GAM fitting was strictly confined to cells within the 1st and 99th percentiles of the DPT distribution. The confidence interval (95\% CI) was computed. Both the original cell image and its corresponding attention map overlay were visualized.

\subsection{scRNA Integration}
\paragraph{Public Data Acquisition and Preprocessing}
Single-cell RNA sequencing (scRNA-seq) data from SNCAX3 and control 3D cortical organoids were obtained from the NCBI Gene Expression Omnibus (GEO) database under accession number GSE236002 \cite{jinModelingLewyBody2024}. The dataset comprised 12 samples (6 controls and 6 SNCAX3) derived from four subclones per genotype, sequenced across three flow cells. One SNCAX3 sample was excluded following the original study's quality-control criteria. Per-sample quality control was performed using Scanpy. Cells expressing fewer than 250 genes or exceeding 25\% mitochondrial transcript fraction were removed. To eliminate putative doublets, cells exceeding both the 99th percentile of detected genes and UMI counts within each sample were excluded. Ambient RNA contamination was corrected per sample using DecontX \cite{yangDecontaminationAmbientRNA2020}, with raw (unfiltered) barcode matrices supplied as background profiles. After concatenation, genes detected in fewer than three cells were removed, and sex-linked genes (Y-chromosome genes, XIST, and TTTY-family transcripts) were excluded to eliminate sex-confounded variation. The 3{,}000 most highly variable genes(HVG) were selected using the variance-stabilizing transformation (Seurat v3 flavour) computed for each batch individually, where each batch was defined as the combination of genotype and subclone identity. Batch correction was performed using scVI with genotype--subclone as the batch key and sex and flow cell as categorical covariates. Mitochondrial and ribosomal transcript fractions were included as continuous covariates. The model was trained for up to 400 epochs with early stopping (patience\,=\,20) using a negative binomial likelihood, yielding a 30-dimensional batch-corrected latent representation used for all downstream analyses. This latent space was subsequently utilized for dimensional reduction and downstreama analysis

\paragraph{Isolation of 2D-Equivalent Neuronal Lineages from 3D Organoid scRNA-seq Data}

To ensure a fair and direct comparison between 2D monolayer cultures and 3D organoids, we specifically subsampled cell populations from the 3D dataset that transcriptomically represent lineages typically yielded in 2D neuronal environments. We evaluated a range of Leiden clustering resolution parameters (0.4, 0.5, and 0.6) based on the scVI latent space, and selected 0.5 as the optimal resolution, as it provided the clearest separation of major cell lineages without over-clustering. 

Cluster-specific marker genes were identified using differential expression gene (DEG) analysis based on raw counts. Based on these canonical markers, we selectively isolated the specific clusters that exhibited transcriptional profiles corresponding to 2D-cultured neurons—primarily comprising "Mature Excitatory Neurons" and "Forebrain Excitatory Neurons"—for subsequent comparative analysis. The detailed DEG profiles defining each cluster are provided in Supplementary Fig. S14.

\paragraph{Gromov-wasserstein Coupling of Image Representations and scRNA-seq data}
Figure 12 provides an overview of the implementation pipeline for GW-map. After the isolation of 2D-Equivalent Neuronal Lineage, image-level SAE representation dimensions were filtered to retain those related to Control and \textit{SNCA}X3-specific feature maps (Wilcoxon rank-sum DE, adjusted $p < 0.05$, $|\log_2\text{FC}| \ge 0.58$). The resulting feature vectors were then log-transformed, $z$-score standardized, and $L_2$-normalized and projected to 30 principal components for Gromov-wasserstein(GW) optimal transport (OT) coupling. On the scRNA side, the 30-dimensional scVI latent representation was used directly without further transformation. Up to 15{,}000 cells per modality were subsampled with genotype-balanced selection (equal Control and SNCA counts).

An entropic GW-OT coupling matrix was computed between the image PCA (d=50) space and the scVI latent space using geodesic intra-modal cost matrices ($k$-NN graph shortest paths; $k = 60$) and Sinkhorn regularization ($\varepsilon = 0.003$). The resulting coupling matrix $\pi \in \mathbb{R}^{n_{\text{img}} \times n_{\text{rna}}}$ defines a soft probabilistic assignment between image and transcriptomic cells. Coupling quality was assessed via genotype matching accuracy (hard assignment and barycentric projection) and GW loss.

Per-gene expression profiles were obtained by applying the trained scVI decoder to yield denoised normalized expression across highly variable genes (3{,}000). For each image cell $i$, a soft gene expression target was constructed via barycentric projection: $\hat{y}i = \sum_j \tilde{\pi}{ij} \cdot g_j$, where $\tilde{\pi}_{ij}$ denotes the row-normalized coupling weight and $g_j$ the gene expression vector of RNA cell $j$.

\begin{figure}[h!]
  \includegraphics[width=1.0\linewidth]{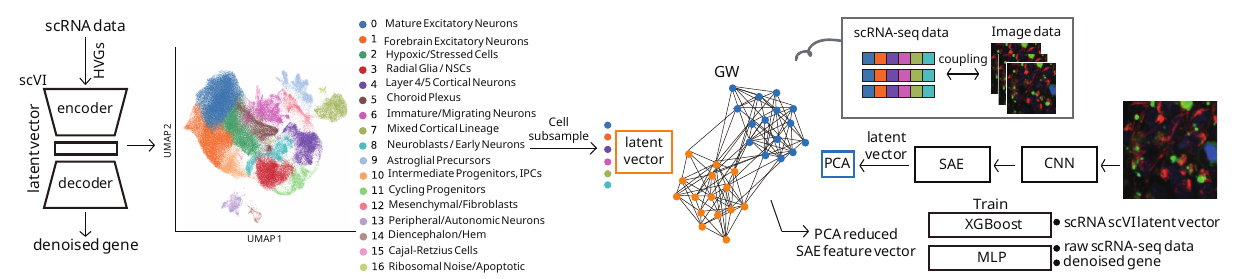}
     \captionof{figure}{Overview of the GW-map implementation pipeline.}
   \label{fig5_overview}
\end{figure}

\paragraph{Prediction of Coupled Image Representations and scRNA-seq Data} 
Two regression models were trained via five-fold stratified cross-validation: XGBoost and a multilayer perceptron(MLP). First, both (i) pre-scVI log-transformed HVG expression and (ii) post-scVI denoised HVG expression were predicted from 100-dimensional image PCA feature vectors via a two-hidden-layer MLP (128 units, dropout 0.4, batch normalization, cosine annealing with learning rate $5 \times 10^{-4}$ and weight decay $1.2 \times 10^{-3}$; early stopping with patience 50, max 200 epochs). MLP inputs and targets were standardized per fold using training-set statistics, and predictions were clipped to the observed training range to prevent numerical instability from sparse genes. Second, (iii) scVI latent vectors were predicted from 30-dimensional image PCA feature vectors via XGBoost (200 estimators, max depth 5, learning rate 0.1). Performance was quantified by $R^2$ averaged across folds. To eliminate genotype as a confounding factor, the coupling matrix was restricted to within-genotype sub-blocks (Control--Control and SNCA--SNCA), and prediction was performed independently within each genotype ensuring that predictive performance reflects genuine cell-level Phenotype--transcriptome correspondence rather than trivial genotype separation.

Coupling rows were randomly permuted to generate null distributions of $R^2$. For per-gene significance, z-scored $R^2$ values were computed from $n = 800$ Ridge (utilized for computational efficiency in large-scale permutation) and $n = 1{,}000$ MLP permutations, enabling to mitigate potential inflation of values due to data distribution or variance structures.

\paragraph{Gene Network Inference Based on Coefficient of Determination}
To identify coherent biological programes among the predicted genes, each gene was represented as a two-dimensional point using its within-genotype permutation $z$-scored $R^2$-one axis for Control and one for SNCA-quantifying how strongly that gene's expression was predicted from phenotypical features within each genotype. A $k$-nearest-neighbour graph ($k = 15$, Euclidean distance) was constructed on these two-dimensional coordinates and community detection was performed using the Leiden algorithm at multiple resolutions (0.002, 0.2, 0.3). Gene flow across resolutions was visualized as a Sankey diagram, allowing hierarchical decomposition of functionally coherent gene modules. Each cluster was annotated via over-representation analysis (ORA; Fisher's exact test, Benjamini--Hochberg adjusted $p < 0.05$) against GO Biological Process, KEGG, Reactome, and MSigDB Hallmark gene sets. The identity of each cluster was subsequently determined based on the most significantly enriched biological terms and the expression of known canonical markers. 

We also selected top 200 high \emph{SNCA x3} z score genes and similarly employed sankey diagram using Leiden algorithm at multiple resolutions (0.08, 0.25, 0.4, 1.0). The identity of each cluster was manually assigned by directly examining the lists of RNA.

Force-directed networks were generated for selected clusters to visualize intra-cluster gene connectivity and sub-cluster organization, with node size proportional to graph degree.

\subsection{CNN Interpretability through the lens of Mechanistic Interpretability}
\paragraph{Quantification of class-specific feature maps across models}
The analysis was coducted with sparsity 3200 instance. The class-specific feature map was selected based on image-level mean activations. Considering that the opposite sign concepts can be superposed to single feature map, raw CNN GAP values were transformed to absolute magnitude. Then identified via differential activation comparing Control to each mutation (sparsity threshold $5\times10^{-5}$; Wilcoxon rank-sum test, adjusted $p < 0.05$; $|\log_2 \text{FC}| > 0.58$).

\paragraph{Visualizing the Feature map via Bilinear Interpolation}

We employ bilinear interpolation to visualize class-specific latent feature maps. For a given input images, the CNN encoder produces a spatial feature map at  ($f_{stage5_out}$). The normalized tokens are passed through the trained SAE encoder, producing an activation vector of the dictionary size per spatial position, excluding dead feature maps.

For each CNN and SAE latent feature map, the resulting activation map ($64 \times 64$) is bilinearly upsampled to the original image resolution ($128 \times 128$) and visualized as a jet-colormap overlay. The activation map is normalized by clipping values below the 50th percentile (median) to zero and mapping the 99.9th percentile to the maximum. The overlay is alpha-blended ($\alpha = 0.5$) onto the Fiji-style linearly scaled microscopy image, where per-channel intensity is linearly mapped from the 10th percentile to the 99.5th percentile into the 8-bit range. The top 20 highest-activating images are displayed for each concept. Concept selection is performed via a differential expression filter.

Additionally, to simulate the CNN's perspective, the input image is processed with per-channel SafeInstanceNormalization. The resulting three-channel tensor is then linearly mapped to 8-bit using a global (cross-channel) 1st to 99th percentile clipping.

\paragraph{Activation Maximization}
To visualize the learned concept of individual SAE feature maps, we employ activation maximization (AM) \cite{erhanVisualizingHigherLayerFeatures2009, olahFeatureVisualization2017a} in Fourier-preconditioned space. For each target feature map, a synthetic 128×128 three-channel input image is optimized to maximally activate the feature map's GAP response through the frozen CNN encoder and SAE.

Optimization is performed in the frequency domain: learnable complex-valued Fourier coefficients are initialized with standard deviation 0.01 and converted to pixel space via inverse FFT weighted by a $1/f^\alpha$ decay mask, which penalizes high-frequency components to encourage naturalistic images. Because the SAE's binary gating produces zero gradients in evaluation mode, we replace the hard gate with a temperature-scaled sigmoid (T = 10), providing differentiable approximation.

The objective function maximizes the target feature map's GAP activation while minimizing L2 decay on Fourier coefficients ($\lambda$ = $1 \times 10^{-4}$) and L1 pixel sparsity ($\lambda$ = $1 \times 10^{-4}$) to promote dark backgrounds. At each optimization step, the pixel-space image is clamped to [-3.5, 3.5] to remain within the distribution seen by the safeinstancenormalized encoder, then subjected to random spatial jitter, rotation ($\pm 15^{\circ}$), and scaling (0.9--1.1) for transformation robustness. Optimization uses Adam with cosine annealing over 2,048 steps at an initial learning rate of 40.0.

To control the visual granularity, AM is performed across a grid of Fourier decay powers ($ \alpha \in $ \{0.4, 0.8, 1.2, 1.6\}), spatial jitter magnitudes (\{4, 8, 12, 16\} pixels), and three random seeds. The resulting synthetic images are visualized per-channel alongside spatial activation heatmaps obtained by bilinear upsampling of the SAE neuron's per-token activation map to the input resolution.

\clearpage

\section{Supplementary Figures}
\subsection{Robustness of Diffusion Pseudotime (DPT) Analysis}
\begin{figure}[h]
    \centering

    \begin{minipage}{0.48\textwidth} \centering \textbf{Control} \end{minipage}
    \hfill
    \begin{minipage}{0.48\textwidth} \centering \textbf{Mutation} \end{minipage}
    
    \vspace{0.5em}

    \begin{subfigure}[b]{0.48\textwidth}
        \centering
        \includegraphics[width=\linewidth]{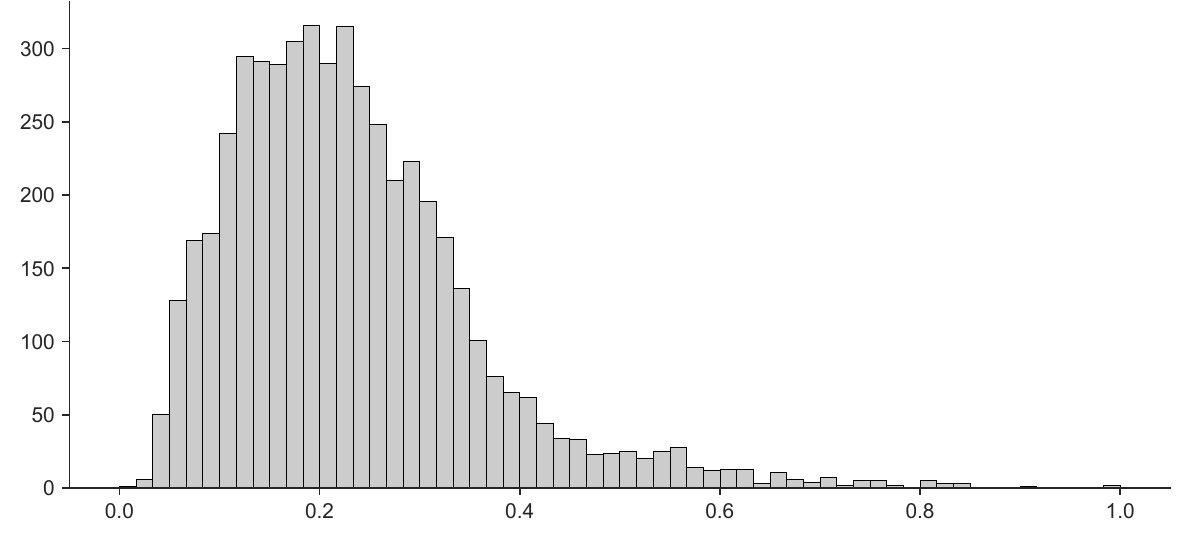}
        \caption*{A. Control (GBA)}
    \end{subfigure}
    \hfill
    \begin{subfigure}[b]{0.48\textwidth}
        \centering
        \includegraphics[width=\linewidth]{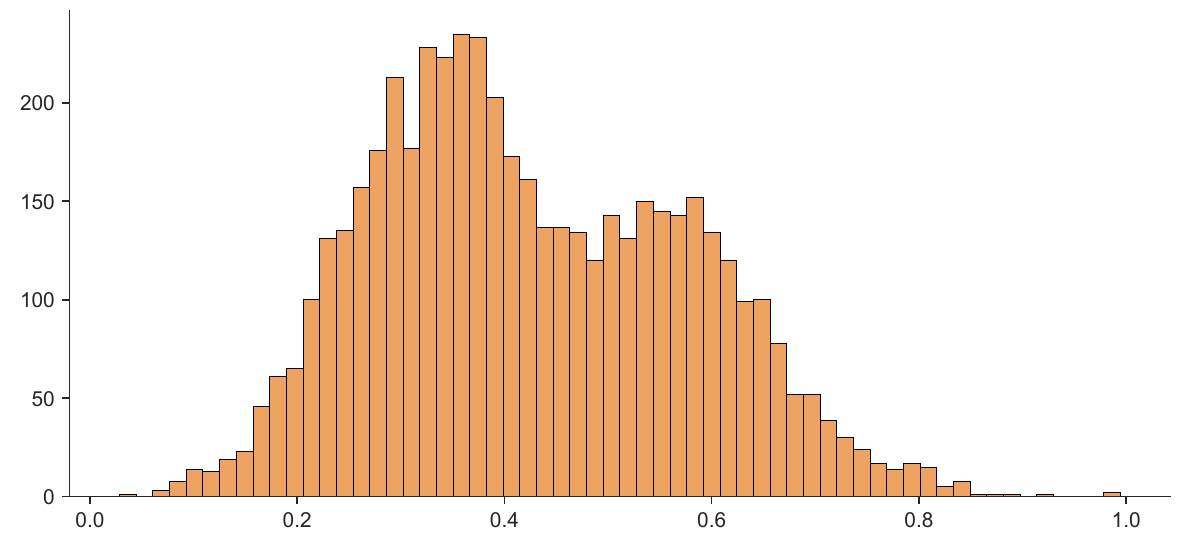}
        \caption*{A. Mutation (GBA)}
    \end{subfigure}

    \vspace{1.5em}

    \begin{subfigure}[b]{0.48\textwidth}
        \centering
        \includegraphics[width=\linewidth]{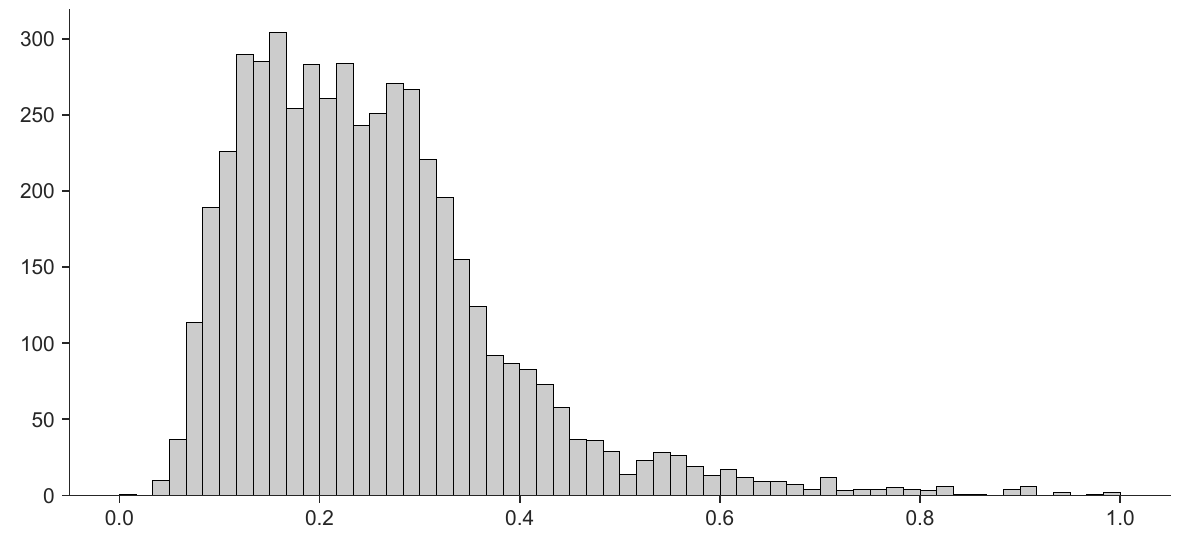}
        \caption*{B. Control (SNCA)}
    \end{subfigure}
    \hfill
    \begin{subfigure}[b]{0.48\textwidth}
        \centering
        \includegraphics[width=\linewidth]{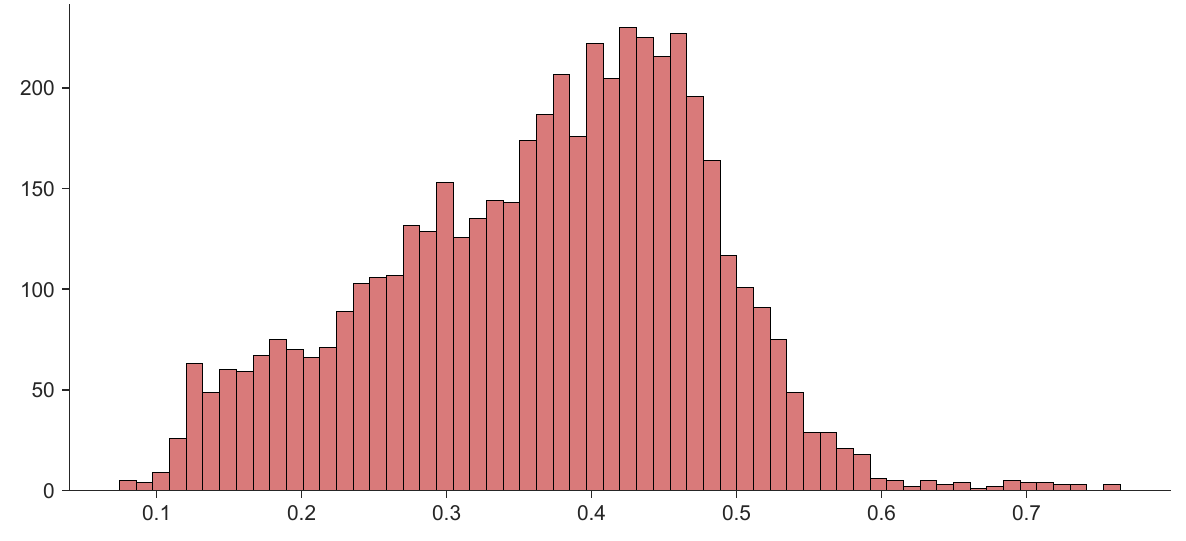}
        \caption*{B. Mutation (SNCAX3)}
    \end{subfigure}

    \vspace{1.5em}

    \begin{subfigure}[b]{0.48\textwidth}
        \centering
        \includegraphics[width=\linewidth]{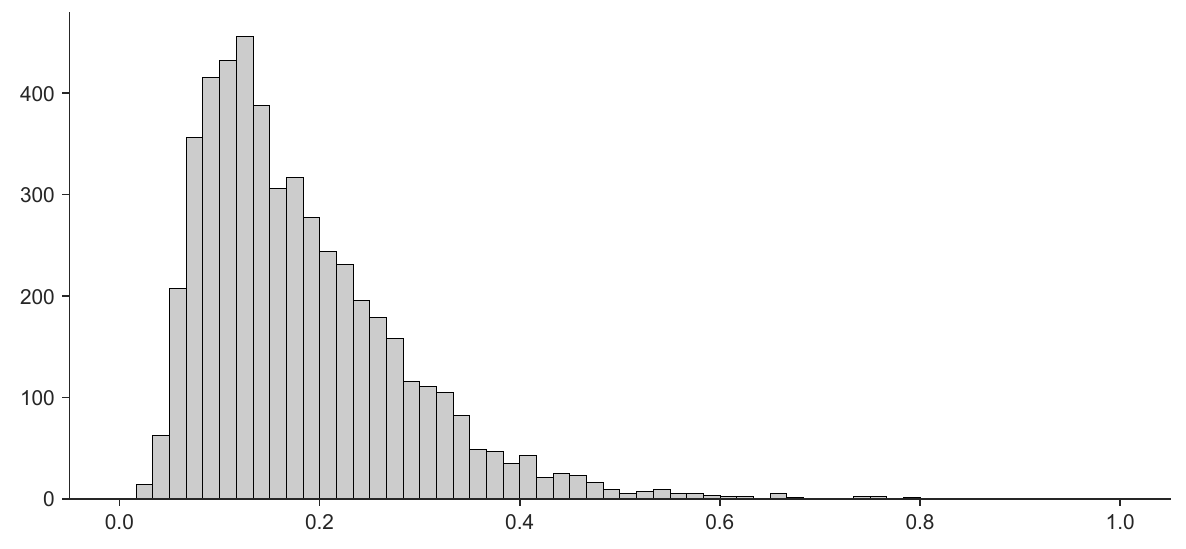}
        \caption*{C. Control (LRRK2)}
    \end{subfigure}
    \hfill
    \begin{subfigure}[b]{0.48\textwidth}
        \centering
        \includegraphics[width=\linewidth]{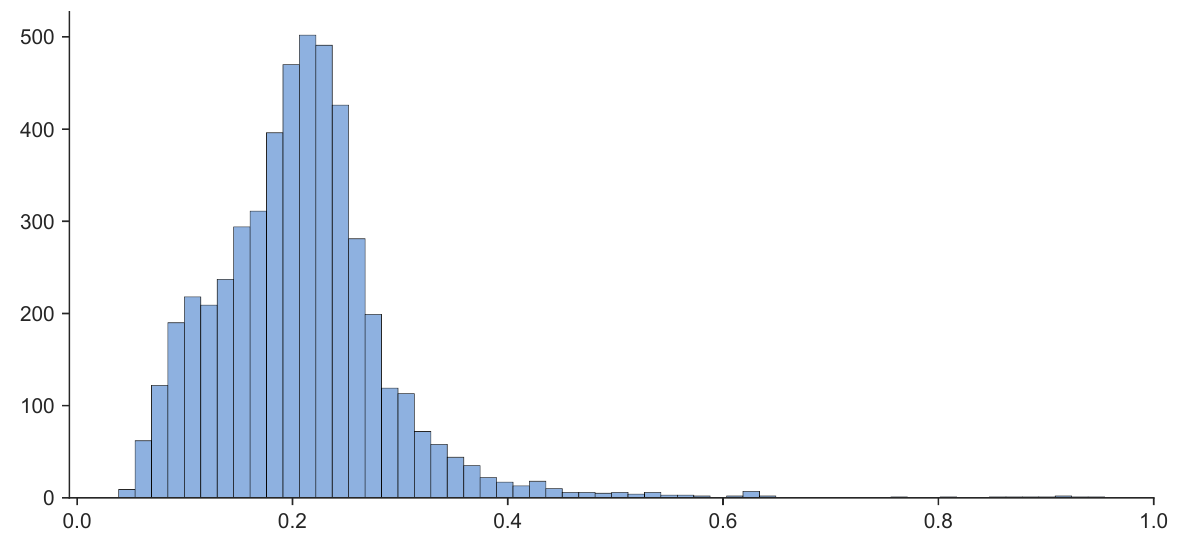}
        \caption*{C. Mutation (LRRK2)}
    \end{subfigure}

    \vspace{1em}
    \caption{\textbf{Distribution of geodesic distance between control medoid and mutation}
    \protect\\ 
    Left column shows control samples, while the right column shows mutation saples (A) GBA, (B) SNCAX3, and (C) LRRK2 lines of geodesic distance (pseudotime) from control medoid.}
\end{figure}

\clearpage

\begin{figure}[ht!]
    \centering

    \includegraphics[width=0.6\linewidth]{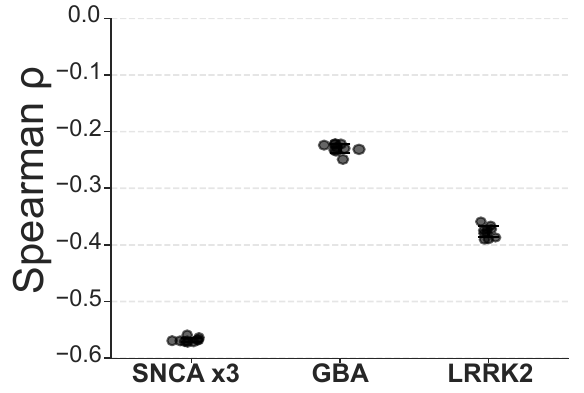}

    \caption{\textbf{Robustness of DPT to Root Cell Selection} \protect\\ 
    To evaluate the stability of our pseudotime estimates, we performed a root permutation analysis by shifting the starting point within the vicinity of the control centroid (n = 10 iterations). For each iteration, the relationship between the distance from the control medoid and the cell death rate within each mutation population was quantified using Spearman’s rank correlation. The consistent distribution of DPT and the stability of these coefficients across permutations demonstrate the robustness of our trajectory model against initial root selection."}

    \label{fig:supp_new_image}
\end{figure}

\clearpage

\subsection{Characterization of Leiden Clusters and Cell-Type Annotation of scRNA-seq data}

\begin{figure}[ht!]
    \centering

    \includegraphics[width=1.0\linewidth]{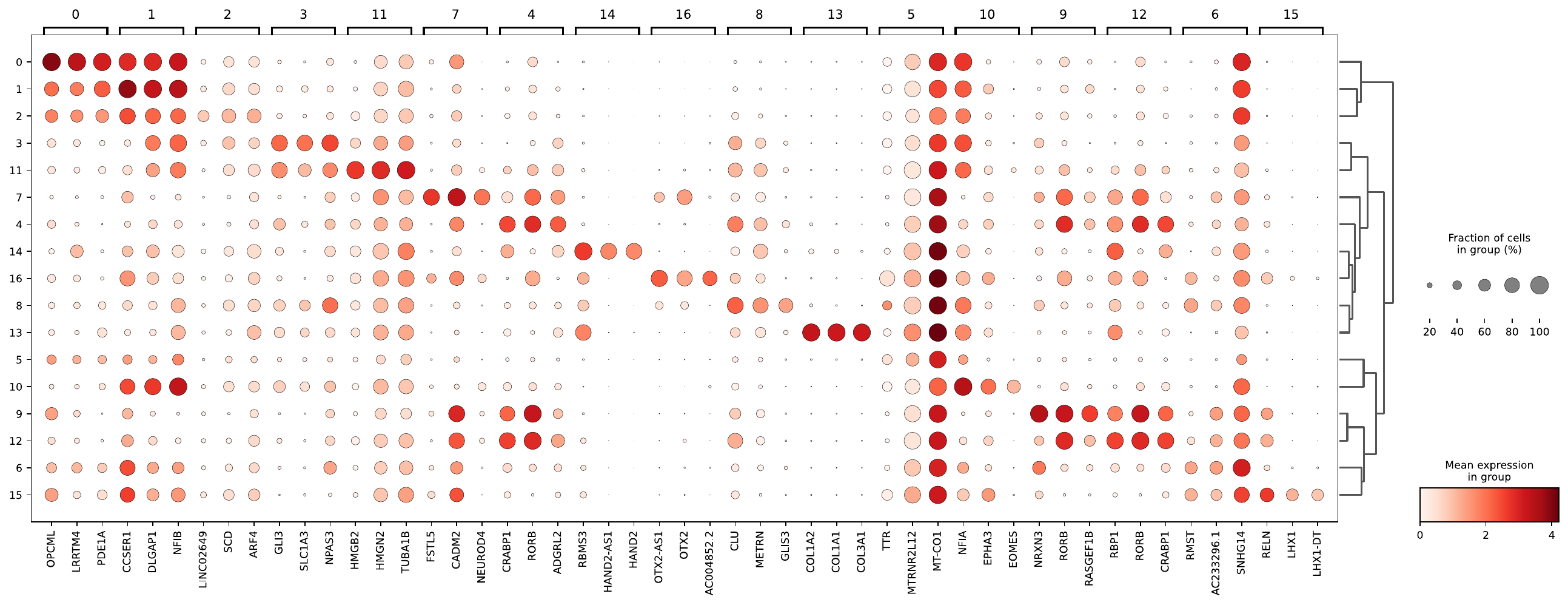}

    \caption{\textbf{Expression patterns of cluster-specific marker genes} \protect\\ 
    Dot plot visualizing the expression of the top 10 differentially expressed genes (DEGs) for each cell cluster. The size of each dot represents the percentage of cells expressing the gene, and the color intensity indicates the average expression level (scaled) within the cluster}
    
\end{figure}

\begin{figure}[p]
    \centering
    \includegraphics[width=1.0\linewidth, trim={0 0 0 0}, clip]{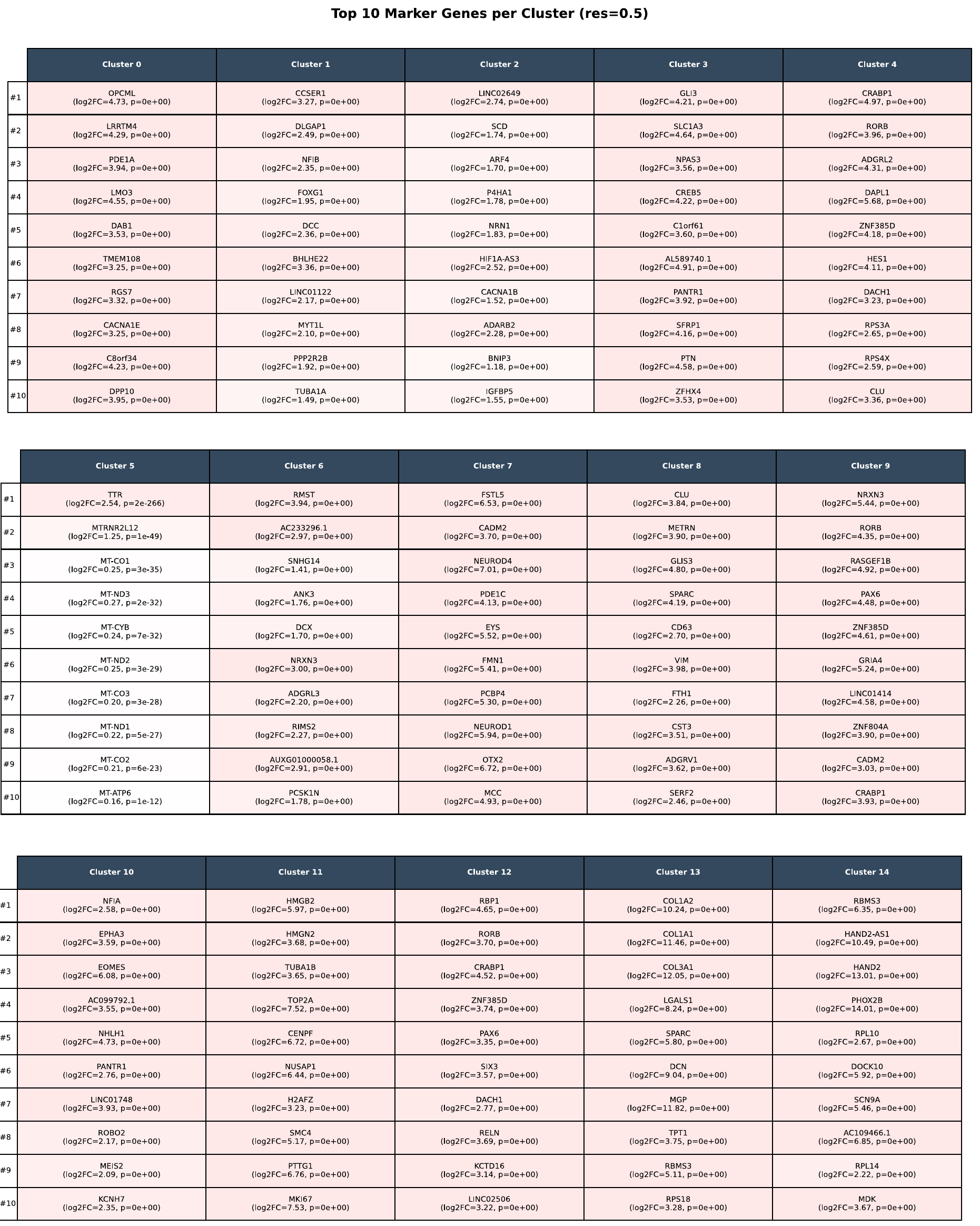}
    \caption{\textbf{Statistical summary of top 10 DEGs} (Continued on next page)}
\end{figure}

\clearpage

\begin{figure}[ht!]
    \ContinuedFloat
    \centering
    \includegraphics[width=1.0\linewidth, trim={0 1300pt 0 0}, clip]{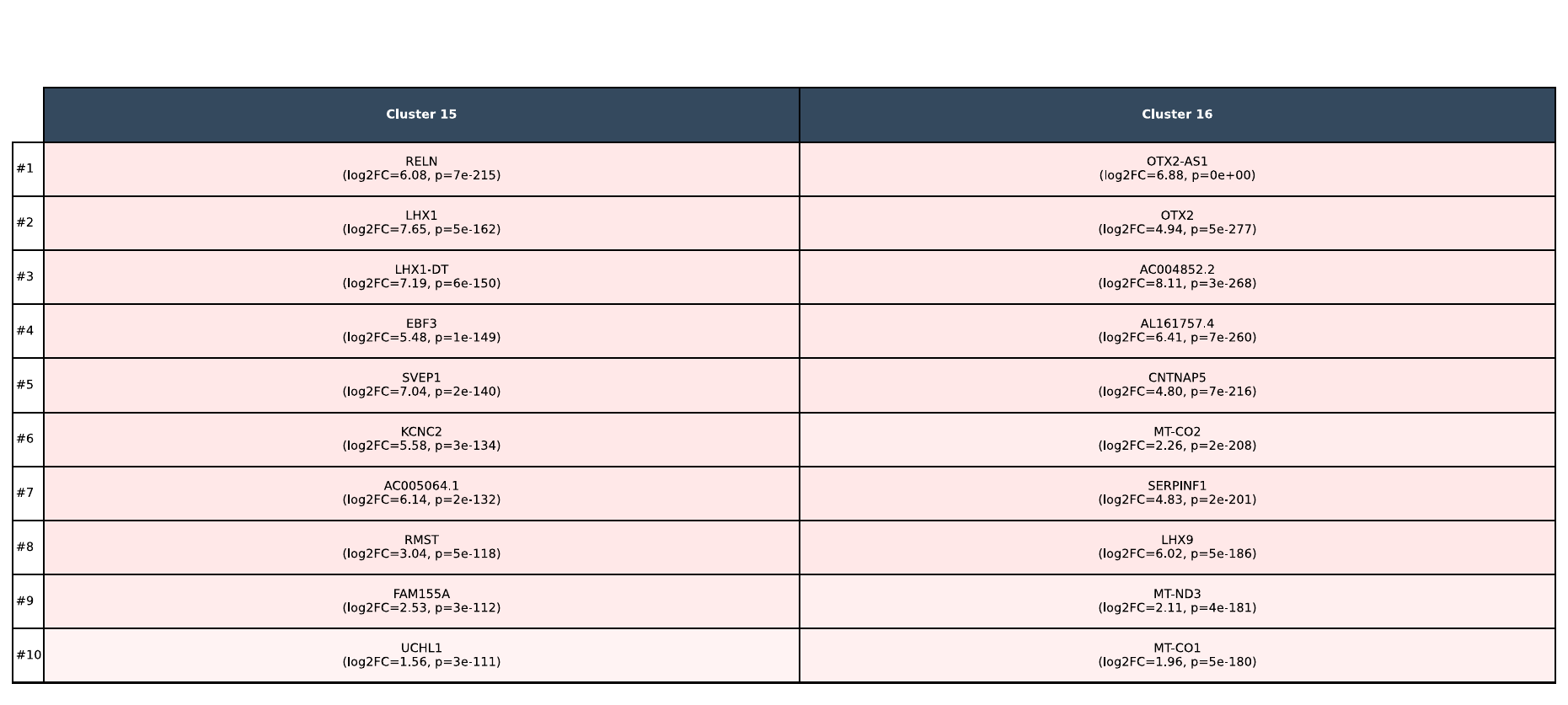}
    
    \vspace{1ex} 
    \caption{\textbf{Statistical summary of top 10 DEGs per cell cluster}
    \protect\\ 
    List of the top 10 most significant differentially expressed genes for each identified cluster. Statistics include gene symbols, average log2 fold-change, and adjusted p-values calculated using the Wilcoxon Rank Sum test.}
    \label{fig:supp_single_example}
\end{figure}

\clearpage

\subsection{Interpretation of Individual Feature Map of SAE (Fig. 17 - Fig. 21)}
Original cell image, attention map cell image, overlay, activation maximization with individual channel, combined and attention map of activation maximization are included. These figures indicate that the individual feature map of SAE is activated by specific concept like mitochondria, mitochondiria-lysosome spatial colocalization, nuclear envelope making each feature map interpretable. This monosemanticity indicates that SAE successfully disentangle the entangled superposition.

\clearpage

\begin{figure}[t]
  \centering
  \nopagebreak
  \includegraphics[width=1.0\linewidth]{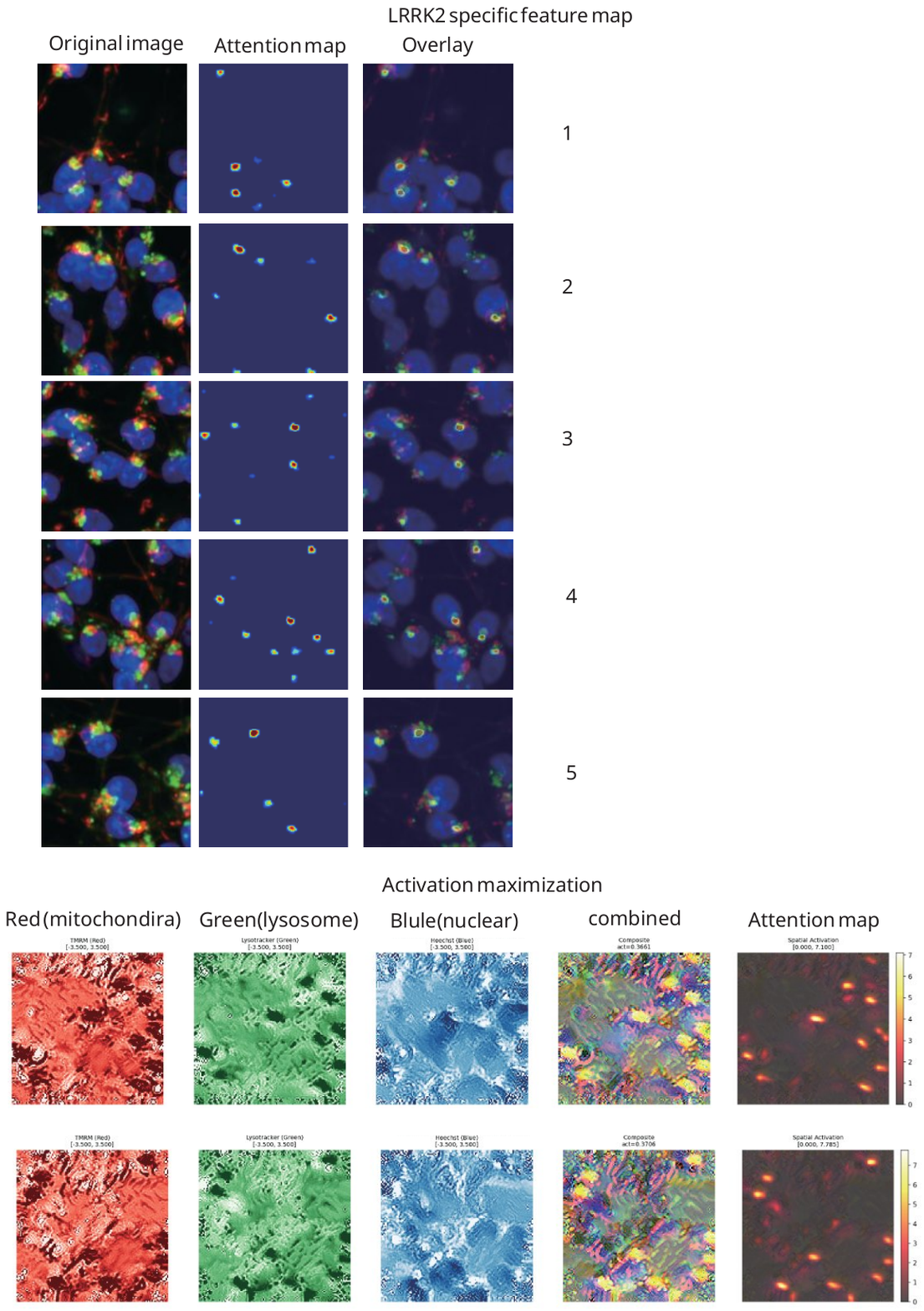}
  \captionof{figure}{Visual illustration of SAE feature map. }
  \vspace{-10pt}
  \label{feature map 0018 LRRK2}
\end{figure}

\begin{figure}[t]
  \centering
  \nopagebreak
  \includegraphics[width=1.0\linewidth]{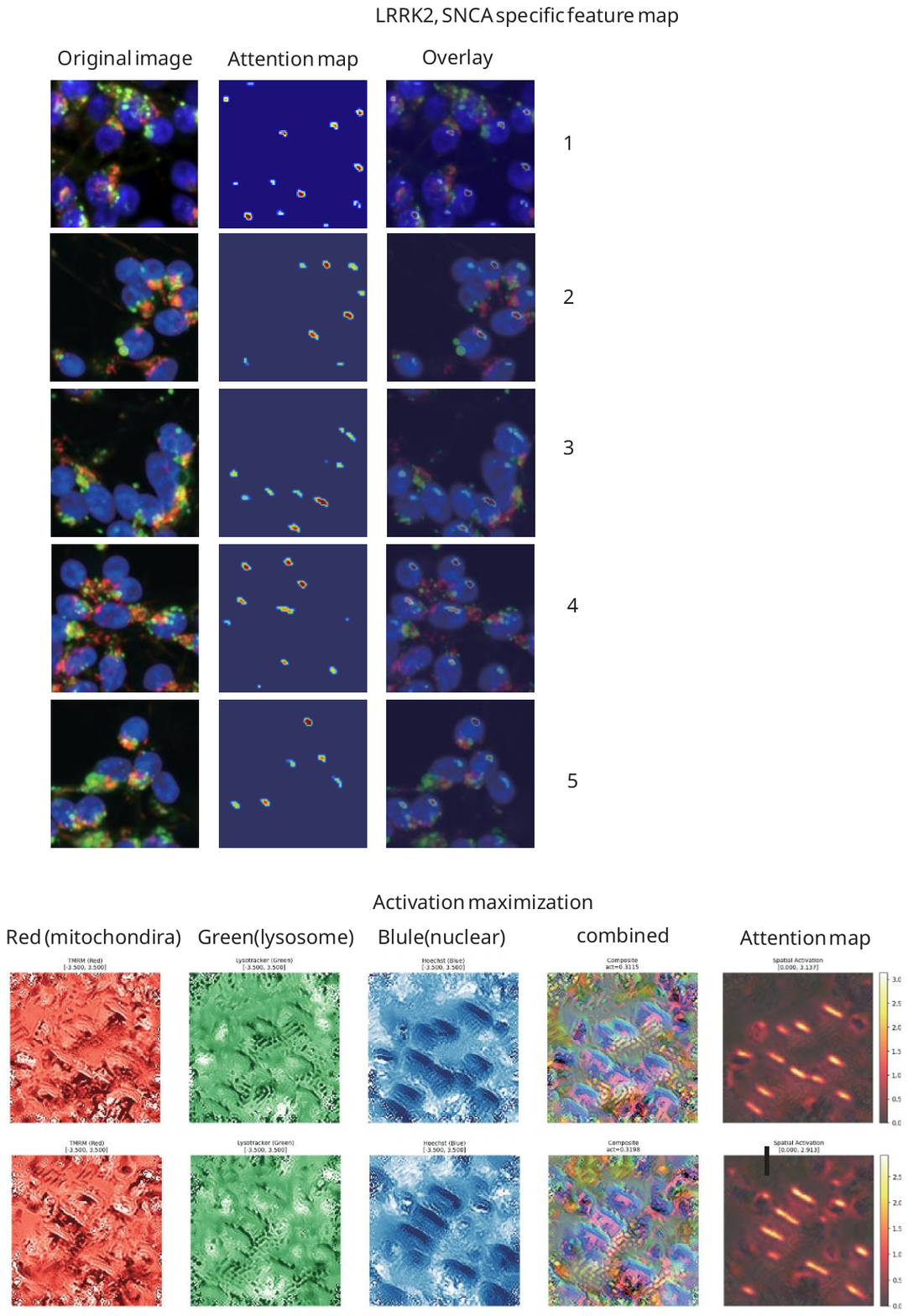}
  \captionof{figure}{Visual illustration of SAE feature map. }
  \vspace{-10pt}
  \label{feature map 0814_LRRK2_SNCA}
\end{figure}

\begin{figure}[t]
  \centering
  \nopagebreak
  \includegraphics[width=1.0\linewidth]{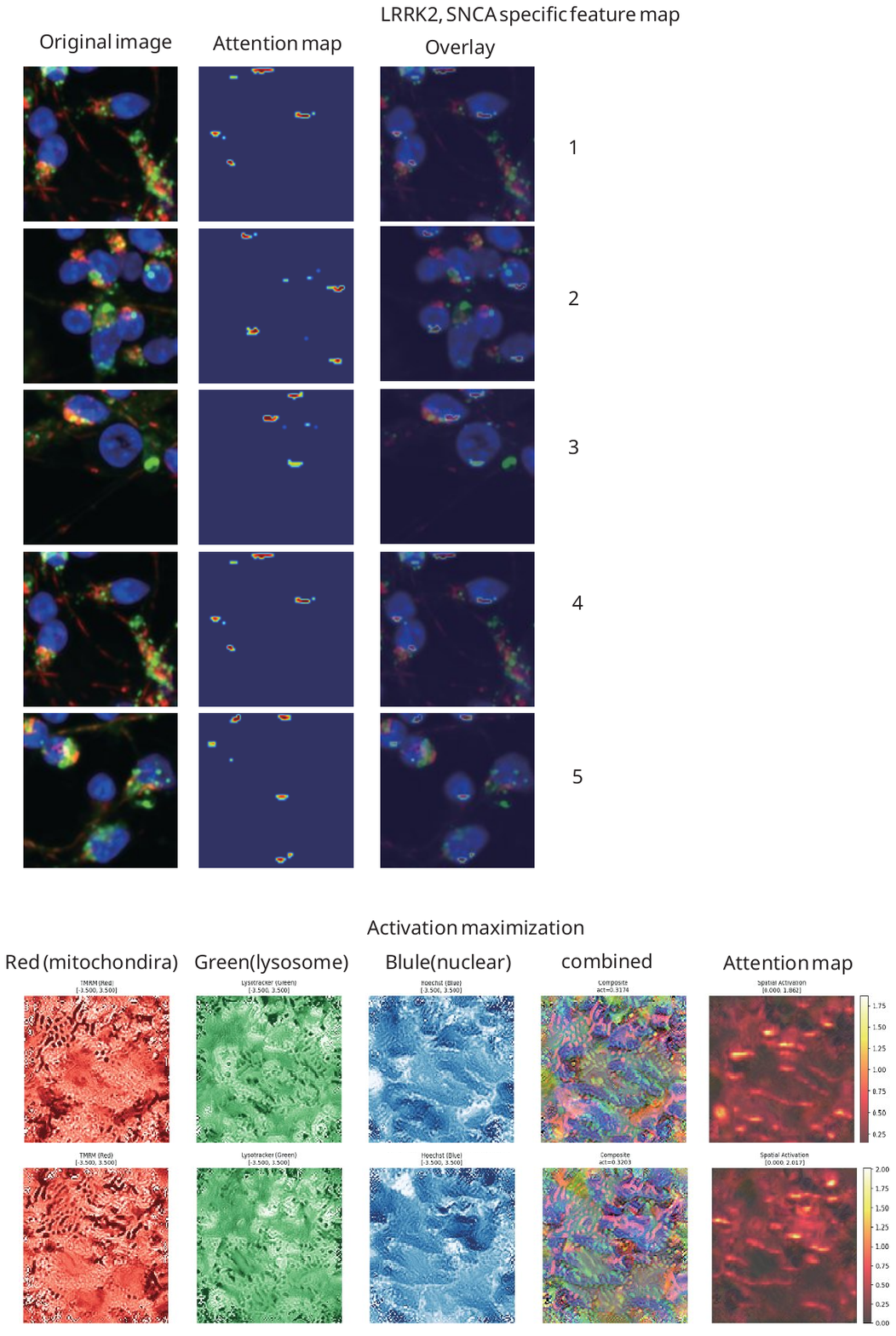}
  \captionof{figure}{Visual illustration of SAE feature map. }
  \vspace{-10pt}
  \label{Feature map 2183 LRRK2_SNCA}
\end{figure}

\begin{figure}[t]
  \centering
  \nopagebreak
  \includegraphics[width=1.0\linewidth]{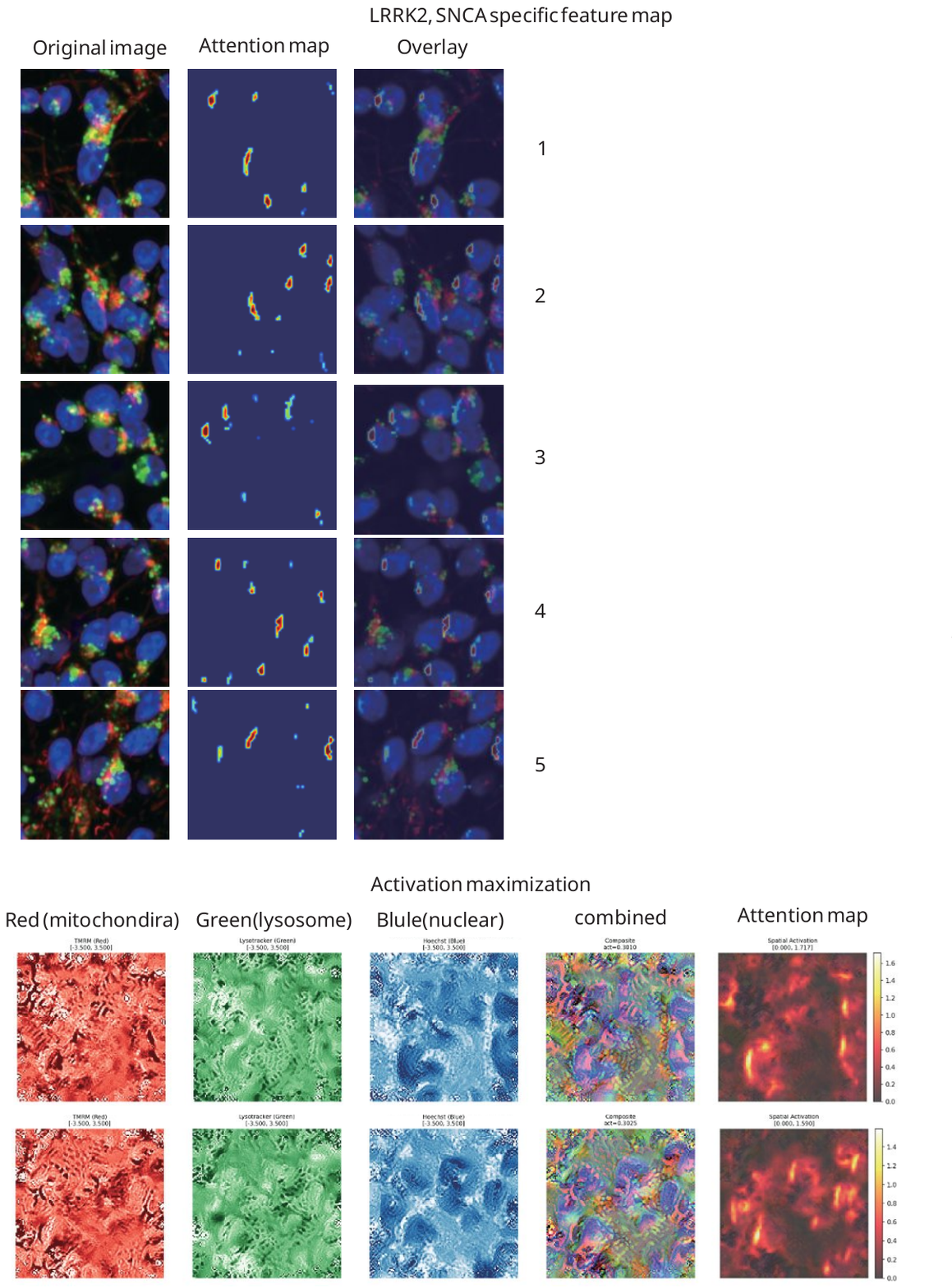}
  \captionof{figure}{Visual illustration of SAE feature map. }
  \vspace{-10pt}
  \label{Feature map 3001 LRRK2_SNCA}
\end{figure}

\begin{figure}[t]
  \centering
  \nopagebreak
  \includegraphics[width=1.0\linewidth]{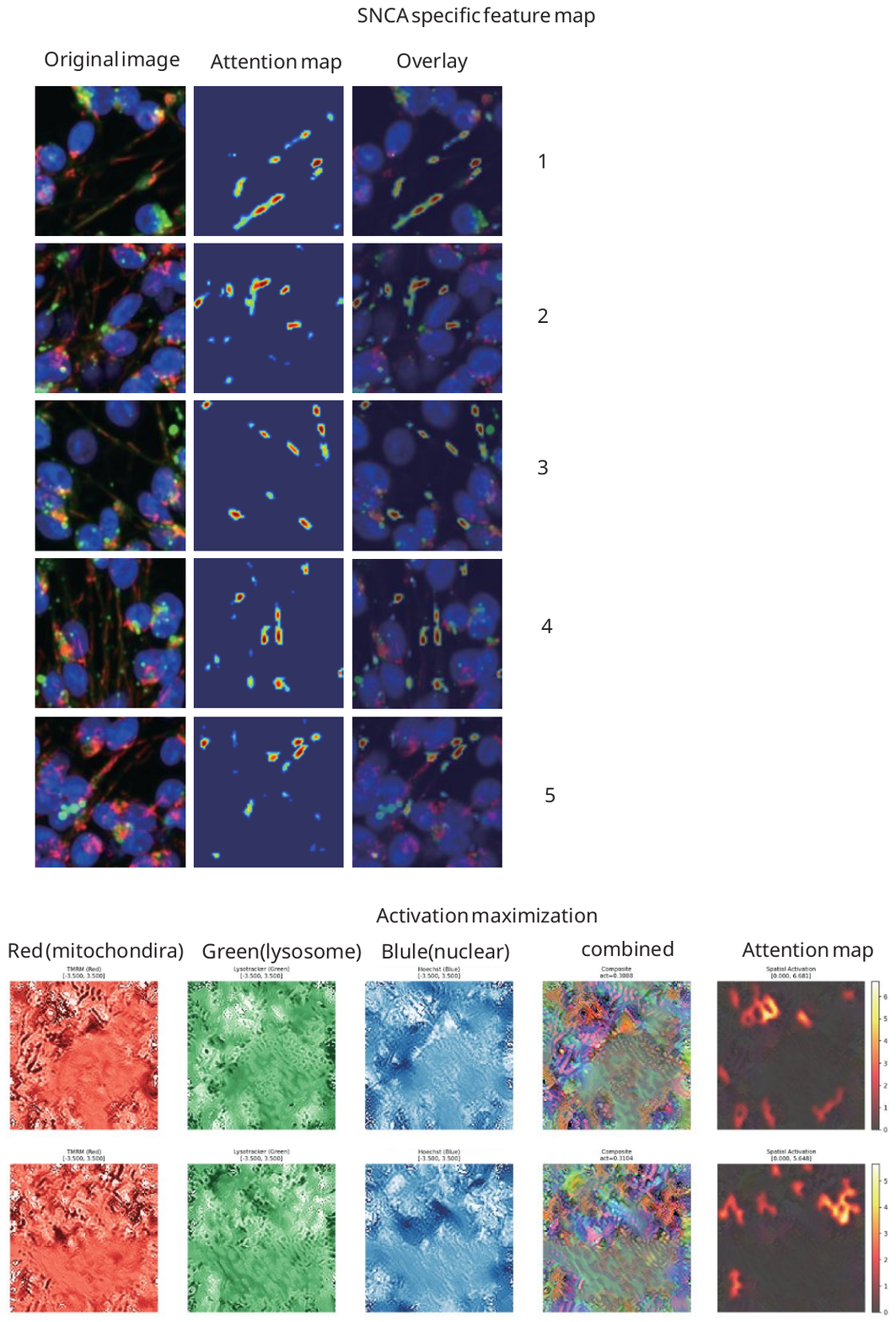}
  \captionof{figure}{Visual illustration of SAE feature map. }
  \vspace{-10pt}
  \label{feature map 1304 SNCA}
\end{figure}

\clearpage

\end{document}